\documentclass[journal]{IEEEtran}
\usepackage{cite}

\ifCLASSINFOpdf
  \usepackage[pdftex]{graphicx}
\else
\fi
\usepackage{algorithmic}
\ifCLASSOPTIONcompsoc
  \usepackage[caption=false,font=normalsize,labelfont=sf,textfont=sf]{subfig}
\else
  \usepackage[caption=false,font=footnotesize]{subfig}
\fi

\usepackage{amssymb}
\usepackage{amsmath}
\usepackage{amsfonts}

\usepackage{url}
\usepackage{multirow}
\usepackage{bm}

\usepackage{enumitem}

\usepackage{color, colortbl}
\usepackage{xcolor}

\definecolor{lightgray}{gray}{0.1}
\definecolor{orange}{rgb}{1,0.5,0}

\usepackage{fancyhdr}
\begin{document}

%

\title{On Trustworthy Decision-Making Process of Human Drivers from the View of Perceptual Uncertainty Reduction}

\author{Huanjie~Wang,~\IEEEmembership{Member,~IEEE,}
        Haibin~Liu,
        Wenshuo~Wang,~\IEEEmembership{Member,~IEEE,}
        and~Lijun~Sun,~\IEEEmembership{Senior~Member,~IEEE}
        
\thanks{This work was supported by the National Key Research and Development Program of China under Grant 2021YFB1716200 and the Canada IVADO Postdoctoral Fellowship Awards. (Corresponding Authors: Haibin Liu; Wenshuo Wang.)} 

\thanks{Huanjie Wang and Haibin Liu are with the College of Intelligent Machinery, Faculty of Materials and Manufacturing, Beijing University of Technology, Beijing 100124, China (e-mail: wanghuanjie@bjut.edu.cn; liuhb@bjut.edu.cn). }

\thanks{Wenshuo Wang and Lijun Sun are with the Department of Civil Engineering, McGill University, Montreal, QC H3A 0C3, Canada (e-mail: wwsbit@gmail.com; lijun.sun@mcgill.ca). }

}


%

\maketitle

\thispagestyle{fancy}


\begin{abstract}

Humans are experts in making decisions for challenging driving tasks with uncertainties. Many efforts have been made to model the decision-making process of human drivers at the behavior level. However, limited studies explain how human drivers actively make reliable sequential decisions to complete interactive driving tasks in an uncertain environment. This paper argues that human drivers intently search for actions to reduce the \textit{uncertainty} of their perception of the environment, i.e., perceptual uncertainty, to a low level that allows them to make a trustworthy decision easily. This paper provides a proof of concept framework to empirically reveal that human drivers' perceptual uncertainty decreases when executing interactive tasks with uncertainties. We first introduce an explainable-artificial intelligence approach (i.e., SHapley Additive exPlanation, SHAP) to determine the salient features on which human drivers make decisions. Then, we use entropy-based measures to quantify the drivers' perceptual changes in these ranked salient features across the decision-making process, reflecting the changes in uncertainties. The validation and verification of our proposed method are conducted in the highway on-ramp merging scenario with congested traffic using the INTERACTION dataset. Experimental results support that human drivers intentionally seek information to reduce their perceptual uncertainties in the number and rank of salient features of their perception of environments to make a trustworthy decision.

\end{abstract}


%
\IEEEpeerreviewmaketitle

\section{Introduction} \label{sec:introduction}
%
%
%
%

\IEEEPARstart{B}{y} continuously interacting with other road users through implicit and explicit communications, humans can complete their driving tasks smoothly in interaction-intensive, safety-critical environments with \textit{uncertainties} \cite{wang2022social}. Rational human drivers negotiate with others by actively perceiving or seeking information about the world via proactively taking specific actions that influence the world. For instance, in real traffic, human drivers usually make tentative attempts before decisively switching lanes to signify their intention of lane changes and tease out the purposes of other vehicles in the adjoining lanes. Providing insights into the underlying decision-making mechanisms of human drivers in uncertain environments can facilitate the development and deployment of autonomous vehicles in challenging scenarios by benefiting advanced algorithms and techniques. 

\begin{figure}[t]
\centering
\includegraphics[width=\linewidth]{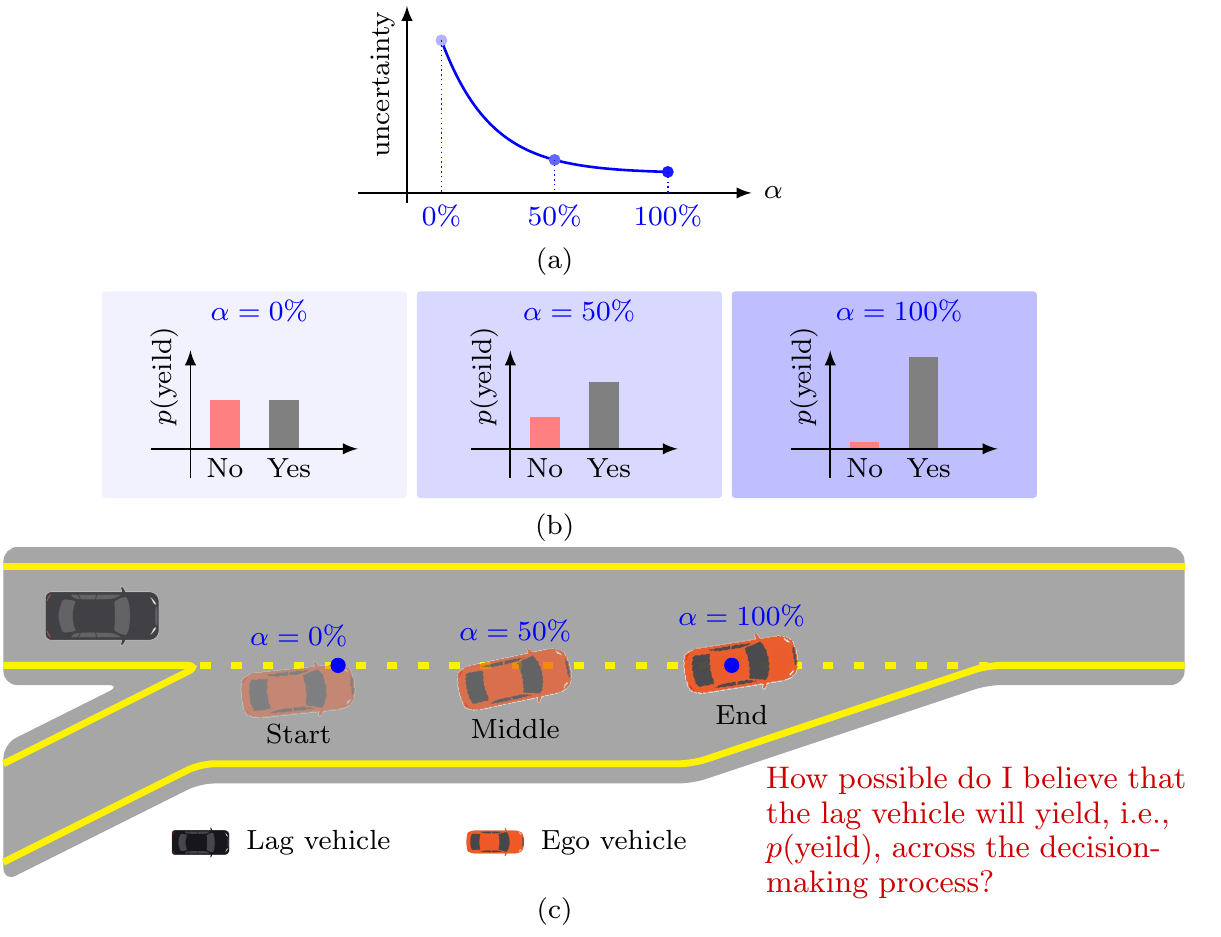}
\caption{Illustration of perceptual uncertainty arising from the ego driver's internal beliefs about the other vehicle's decisions/intentions during decision-making. For more detailed definitions of the scenario, refer to Section III.}
\label{fig:freeway merge example}
\end{figure}

\subsection{Where Uncertainties Come From?}
In the real world, uncertainty is a pervasive part of human interaction and arises when humans lack confidence in their ability to predict the future and explain past results \cite{berger1982language, berger1974some}. In traffic, the surrounding environment is not fully observable to human drivers, and other vehicles' intentions are usually implicit \cite{li2022pomdp}. People have difficulty predicting future outcomes when they are unsure which possible case is most likely to happen. Taking the freeway merge task in Fig. \ref{fig:freeway merge example}(a) for example, the driver in the red car cannot directly read the lag vehicle's intent to yield or not but instead estimates and updates their \textit{beliefs} about how possible it is that the lag vehicle will yield based on qualitative observations of speed, relative speed, and distance between them. Therefore, uncertainties are subjective and correlated to human drivers' knowledge of the human environment and thus can evaluate people's confidence in their prediction and explanatory abilities. In general, uncertainty is related to the number of outcomes likely to happen \cite{shannon1948mathematical, schubert2011evaluating}. Uncertainty is low when there is only one choice and high when several alternatives are equally likely to happen. Moreover, uncertainty also depends on human beliefs about the outcomes of these alternatives, which are usually characterized by probability. For example, the driver in the red vehicle has the highest uncertainty about the intent of the black vehicle if the driver's probabilistic beliefs about the yield or not are equal, i.e., $p(\mathrm{yield} = \mathrm{Yes}) = p(\mathrm{yield} = \mathrm{No}) = 0.5$ when making decisions at the initial stage $\alpha=0\%$, as shown in the left plot of Fig. \ref{fig:freeway merge example}(b). This is factual since it is hard for human drivers to decide to merge or yield safely when unsure about the other vehicle's intentions or decisions. On the contrary, when the merge behavior continues, the driver in the red vehicle can get more practical information to update and increase their beliefs about the black vehicle's intention, as shown in the right plot of Fig. \ref{fig:freeway merge example}(b) with $\alpha=100\%$. The uncertainty of their beliefs about other vehicles' intentions across the entire decision-making process would reduce as schematized in Fig. \ref{fig:freeway merge example}(a).

The above analysis, with an example of merging behavior in human environments, indicates that uncertainty mainly stems from two factors: (i) the number of possible outcomes and (ii) the probabilistic distributions of these outcomes. Researchers usually describe the possible outcomes as a random feature to quantify the perceptual uncertainty and then evaluate the uncertainty using information theory such as entropy \cite{molina2021robotic}. This paper will mainly focus on the relationship between perceptual uncertainty and the distributions of possible outcomes during the decision-making process.

\subsection{Uncertainty in Human Decision-Making}
Theoretically, human agents should actively seek helpful information to update their beliefs about the environment and reduce their perceptual uncertainty to make trustworthy decisions \cite{hubmann2018automated, roitberg2022my}. In other words, rational human drivers would not risk themselves by making unreliable decisions when the information about the possible outcomes is insufficient. For instance, the driver in the red vehicle would not blindly take a merge-in action when their beliefs $p(\mathrm{yield})$ are characterized by the left plot of Fig. \ref{fig:freeway merge example} (b), which results in the highest entropy value from the perspective of information theory, i.e., the highest uncertainty. However, the driver would be more confident in merging if they had the beliefs shown by the right plot of Fig. \ref{fig:freeway merge example} (b). Therefore, we argue that the perceptual uncertainty would continuously decrease until a threshold allows the driver to make trustworthy decisions. This paper will empirically develop a proof of concept framework based on an open dataset to support this claim.

\subsection{Contributions}
\begin{figure}[t]
\centering
\includegraphics[width=0.9\linewidth]{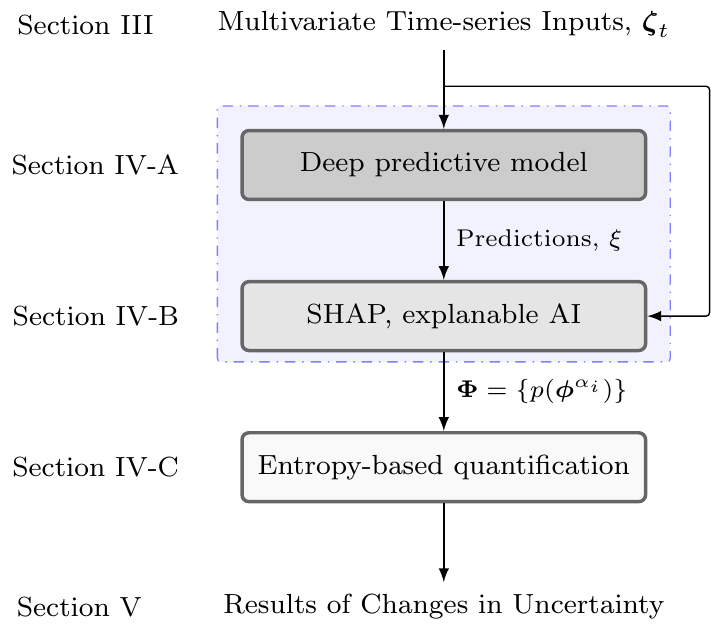}
\caption{Our proposed framework of quantifying perceptual uncertainty of human drivers during decision-making.}
\label{fig:framework}
\end{figure}

Given a new driving task in human environments, we hold that human drivers have uncertainty about the environment since only partial information is observable and other human agents' behaviors are probabilistically multi-modal. Fig. \ref{fig:framework} illustrates our designed framework by integrating with explainable Artificial Intelligence (XAI) and information theory to empirically reveal that perceptual uncertainty decreases when merging on the highway in congested traffic. Our contributions are threefold:
\begin{itemize}
\item Developing a proof of concept framework to investigate the changes in perceptual uncertainty when making decisions in interactive driving tasks.
\item Introducing an XAI approach and combining it with a deep learning-based prediction model to infer the salient features for human drivers to make decisions. 
\item Providing an entropy-based quantification approach to analyze human drivers' perceptual uncertainty changes.
\end{itemize}

\subsection{Paper Organization}
The remainder of this paper is organized as follows. Section II discusses related work on time-series data prediction and explainable methods. Section III introduces the specific scenario and data extraction. Section IV introduces the deep learning-based prediction model, an explainable approach (i.e., SHAP), and entropy-based evaluation methods. Section V analyzes the experimental results and provides further discussion. Section VI gives the conclusion.

\section{Related Work}

\subsection{Time-Series Data Prediction with AI Models}
The input features of AI predictive models can be non-time series or time series. Time series data is measured during a specific period and is dominant in human driver behavior prediction \cite{takano2008recognition}. Traditional classical time-series prediction algorithms include Auto-Regressive Integrated Moving Averages (ARIMA), and exponential smoothing (e.g., Holt-Winters). Although these methods are robust, they usually perform poorly in generalization and have limited learning freedom. Time-series data prediction can be formulated as a supervised learning problem, which accurately maps the input features to the output targets \cite{saluja2021towards}. AI predictive models can offer superior methods for spotting systematic patterns and integrating various sources of information relevant to forecasting processes \cite{hogarth1981forecasting, barnes1984cognitive, corredor2014cognitive, chen2020driving}. Traditional machine learning-based methods of time-series data prediction are $k$-nearest neighbor, support vector machine, random forest, XGBoost, and LightGBM. However, a satisfying predictive model usually requires retaining the structurally spatiotemporal information embedded in the data sequence, which is highly time-consuming and depends on expert experience. Fortunately, deep learning-based predictive models can achieve a high degree of freedom in developing and training through an end-to-end framework. The Long Short-Term Memory (LSTM) network is popular, which is optimized by unrolling the neural network and back-propagating the errors through the entire input sequence. LSTM enables the network to benefit from the temporal relationships among the data samples. Therefore, we developed an LSTM-based network to make predictions in this paper (see Section \ref{subsec:LSTM}).

\subsection{Explanation of Learning-Based Predictive Models}

Deep learning-based predictive models can capture correlations among samples. However, it is also essential to understand the reasons behind their predictions and improve the predictive performance \cite{doshi2017towards}. Besides, getting a model deployed without disclosing the reason for its predictions may lead to inaccurate or potentially dangerous decisions. In general, explainability is the degree of understanding of the reasons for predicted results \cite{miller2019explanation, omeiza2021explanations}. Much research on the XAI \cite{arrieta2020explainable} has been done on their \textit{complexity}, \textit{scope} (local or global), and \textit{specific/agnostic models} \cite{adadi2018peeking}.

\subsubsection{Complexity}
Explainability can be achieved by constructing models with low-complexity structures with inherent explainability, such as logistic regression, linear regression, and decision trees. However, simple models are usually inferior to complex ones \cite{adadi2018peeking, loyola2019black}, thereby increasing inherent explainability usually at the cost of prediction performance \cite{ozyegen2022evaluation}. Interpreting machine learning models with post-hoc explainability after model training is an alternative method by introducing perturbations into the input and observing the resulting model outputs. The explainability is then obtained by reverse engineering the prediction and the model internals.

\subsubsection{Scope}
Explanations can be local or global according to whether they target a single forecast or the whole model. Local explainability focuses on the impact caused by the input features on the model output of specific instances. In contrast, global explainability aims to understand the overall impact of the input features on the model output \cite{molnar2020interpretable}. Hence, local explainability has been widely used due to its flexibility in development. Still, it is challenging to determine the black-box model's overall logic.

\subsubsection{Specific/Agnostic models}
Model-specific tools can be computationally efficient, but only for specific and intrinsically explainable models. In contrast, model-agnostic tools can overcome the trade-off between performance and explainability by ignoring the internal information of the model and not extracting the complex relationship between the inputs and the target. As a result, they are usually preferred when comparing the explainability of various AI models \cite{mujkanovic2019explaining}.

Assessing the feature saliency for decisions can explain data and models \cite{marcilio2020explanations, lim2021temporal}, such as the impacts of time-series features on predictions \cite{guo2019exploring}. 
Many XAI techniques are developed for local post-hoc explanations \cite{das2020opportunities, angelov2021explainable}, such as SHAP \cite{lundberg2017unified}, DeepLIFT \cite{shrikumar2017learning}, Anchors \cite{ribeiro2018anchors}, LoRE \cite{guidotti2018local}, LIME \cite{ribeiro2016should} and its variants \cite{hall2017machine, sokol2020limetree, zafar2021deterministic}. These methods employ surrogate models for each prediction sample with additional perturbations to the features and learn the behavior of the reference model in the particular case of interest. Thus, they can estimate the local feature importance \cite{arrieta2020explainable}. Existing works show that the SHAP technique is more reliable and precise for time-series data than others since it considers \textit{all} potential combinations of input features with or without the presence of each feature \cite{schlegel2019towards, saluja2021towards, ozyegen2022evaluation}. SHAP is a feature attribution method and only needs to obtain a surrogate function of the deep learning-based prediction model to present its explanation. Moreover, SHAP is a unified local explanation approach and thus can ensure the prediction of a fair distribution among input features based on cooperative game theory \cite{lundberg2017unified}. Therefore, we employ the SHAP method in this work to explain our deep learning-based predictive model, see Section \ref{subsec:SHAP}.

\section{Dataset and Data Processing}

\subsection{Scenario Selection and Real-World Dataset} 
\label{subsec:dataset}
\subsubsection{Scenario Selection} 
\begin{figure}[t]
\centering
\subfloat[Real scene]{\label{level1.sub.1} \includegraphics[width=0.98\linewidth]{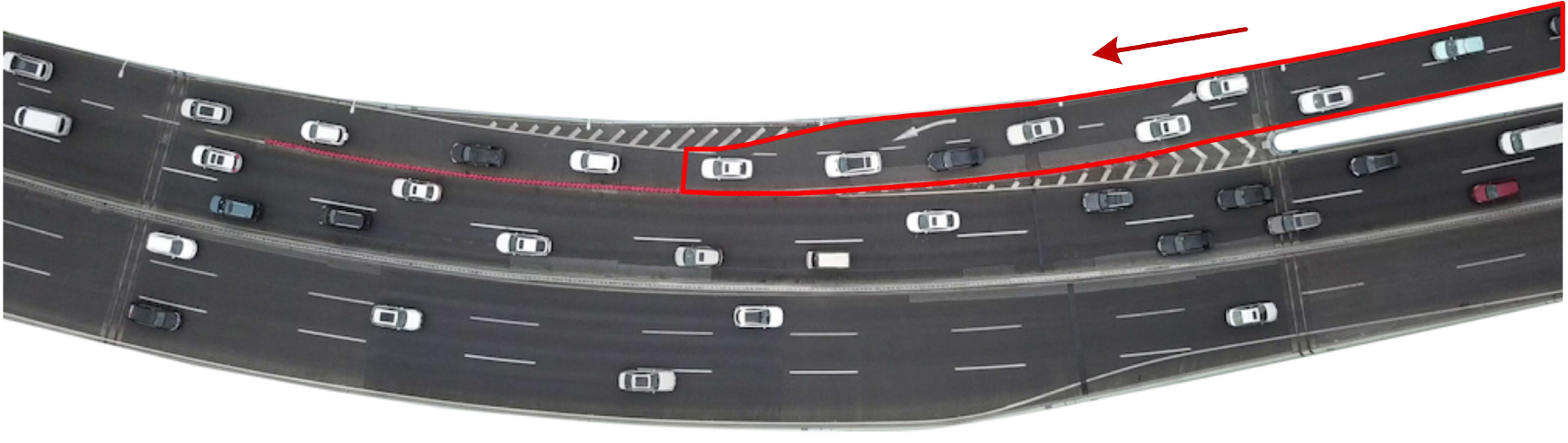}} \\
\subfloat[Data visualization]{\label{level1.sub.2} \includegraphics[width=0.98\linewidth]{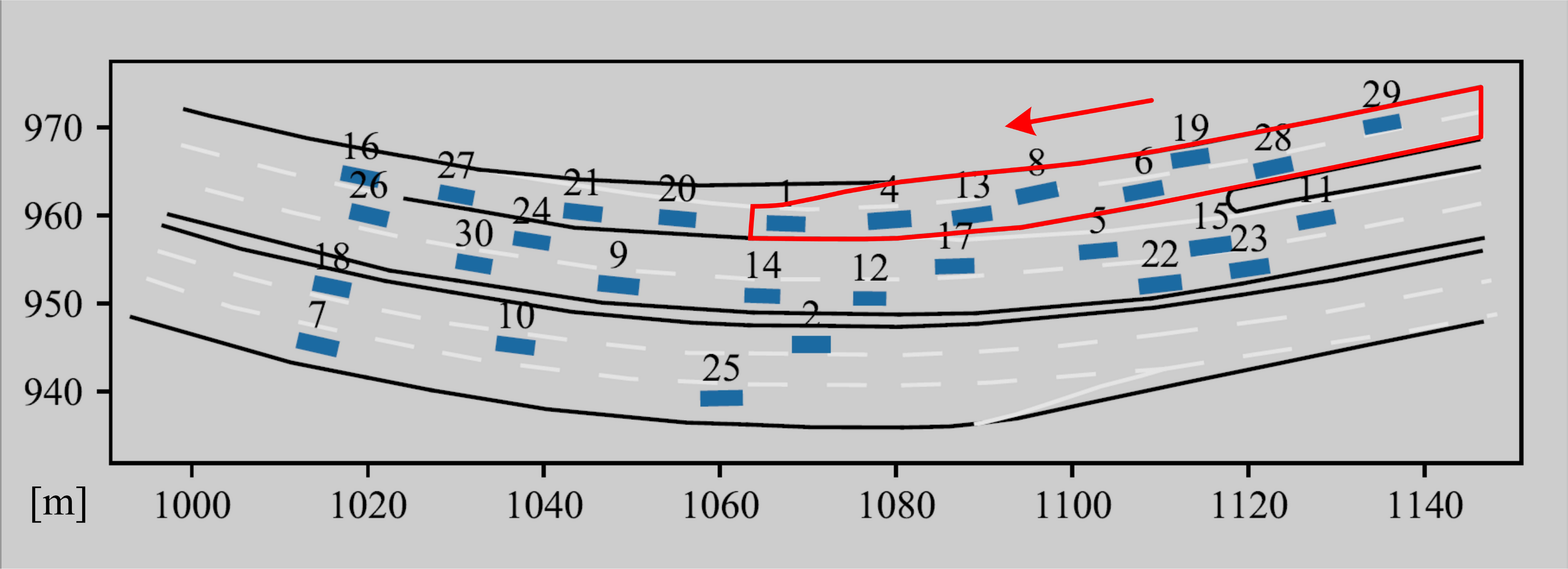}}
\caption{The Chinese highway on-ramp merge scenario in the INTERACTION dataset \cite{zhan2019interaction} and the selected local region rounded by a red line.}
\label{fig:INTERACTION dataset}
\end{figure}

Many open data sets cover diverse traffic scenarios that benefit from advanced sensing equipment (e.g., drones and cameras) and data processing techniques. In this paper, we prefer interaction-intensive and safety-critical scenarios in which uncertainty is integral to human interaction. Therefore, we selected the highway on-ramp merge scenario in congested traffic without traffic control, as shown in Fig. \ref{fig:INTERACTION dataset}(a). When merging on the highway, the human driver needs to take associated actions to seek more information to increase their \textit{confidence} or \textit{beliefs} about other surrounding vehicles' underlying purposes, such as yielding or not. 

\subsubsection{Dataset} 
This paper uses the INTERACTION dataset \cite{zhan2019interaction} collected from the real world by considering the diversity in scenarios and behaviors and the clarity of definitions.
\begin{itemize}
\item \textbf{Scenario diversity:} This dataset covers interaction-intensive driving scenarios (e.g., highway on-ramp merge) collected from China. Some samples were also collected from other countries (i.e., the United States and Germany), but their traffic speeds are fast, and the distribution of vehicles over space is sparse.
\item \textbf{Behavior diversity:} This dataset consists of regular, safe, and interactive driving behaviors such as adversarial, irrational, and near-collision maneuvers. 
\item \textbf{Clear definition:} This dataset has well-defined physical variables such as position, speed, yaw angle, dimension, and type of vehicle, requiring low effort to preprocess data. 
\end{itemize}

\subsection{Data Processing} \label{subsec:data processing}
The total video length of the Chinese highway on-ramp merge scenario in the INTERACTION dataset is $94.62$ minutes with $10,359$ vehicles. Through visualization and analysis of the data, we found that the scenario in the upper two lanes has a high traffic flow and covers diverse driving behaviors, as marked in the red-bounded area in Fig. \ref{fig:INTERACTION dataset}.

\begin{figure}[t]
\centering
\includegraphics[width=\linewidth]{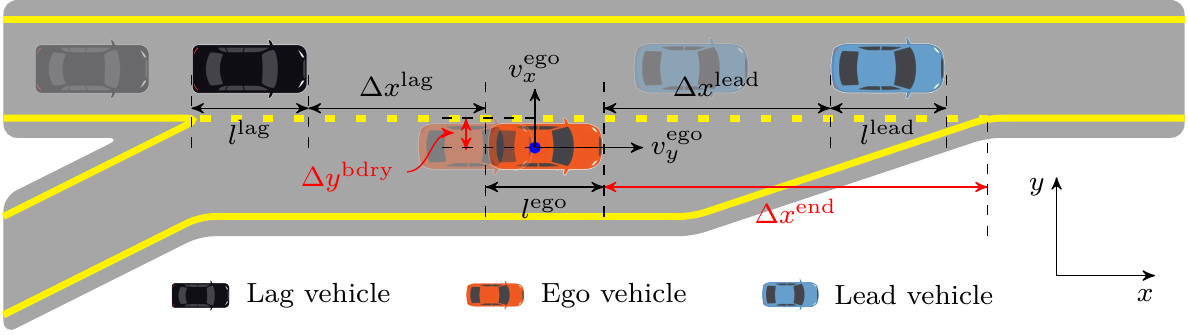}
\caption{Definition of features for the highway on-ramp merge scenario.}
\label{fig:Diagram of a typical highway on-ramp merge scenario}
\end{figure}

\begin{table}
\centering
\caption{Definition of Decision Features}
\label{tab:definition}
\begin{tabular}{l l p{0.6\columnwidth}}
\hline\hline
Symbol & Unit & Definition \\
\hline

    $\Delta {x}^{\mathrm{lead}}$ & [m] & The longitudinal distance between the lead and ego vehicles \\
		
	$\Delta v_{x}^{\mathrm{lead}}$ & [m/s] & The longitudinal speed difference between the lead and ego vehicles \\

	$v_x^{\mathrm{ego}}$ & [m/s] & The longitudinal speed of the ego vehicle \\

	$v_y^{\mathrm{ego}}$ & [m/s] & The lateral speed of the ego vehicle \\

	$\Delta {x}^{\mathrm{lag}}$ & [m] & The longitudinal distance between the lag and ego vehicles \\

	$\Delta v_{x}^{\mathrm{lag}}$ & [m/s] & The longitudinal speed difference between the lag and ego vehicles \\
	
	$\Delta {x}^{\mathrm{end}}$ & [m] & The longitudinal distance of the ego vehicle to the end of the ramp \\
	
    $\Delta {y}^{\mathrm{bdry}}$ & [m] & The lateral distance to the lane change boundary line \\

\hline\hline
\end{tabular}
\end{table}

To define agents conveniently, we simplify the merge scenarios as Fig. \ref{fig:Diagram of a typical highway on-ramp merge scenario} to represent the actual scenario in Fig. \ref{fig:INTERACTION dataset}. The red vehicle going to merge on the highway is denoted as the ego vehicle. The lag vehicle is the black vehicle on the highway and behind the ego vehicle. The blue vehicle on the highway and in front is the lead vehicle. Table \ref{tab:definition} lists some well-defined potential features, and Fig. \ref{fig:freeway merge example} illustrates three critical moments of the decision-making process (i.e., $\alpha = 0\%$, $\alpha = 50\%$, and $\alpha = 100\%$). Note that we define the decision-making endpoint as when the ego vehicle's center crosses the lane boundary rather than the center of the highway lane. This is because our focus is mainly on the decision-making process rather than the entire merge task. When the ego vehicle's center is located at the lane boundary, it is reasonable to assume that the human driver has sufficient confidence to take a merging action safely. 

We finally extracted 796 demonstrations in total ($D=796$), which were then randomly divided into a training set (80\%) and a test set (20\%). In addition, to reduce the influence of the variance in the length of merge demonstrations, we filtered the extracted data using down-sampling/up-sampling techniques and then took realignment.

\section{Methodology}
This section first introduces the proposed XAI framework consisting of the deep learning-based predictive model and the explanation module for the prediction results. The explanation is to evaluate and rank the feature contribution (i.e., feature saliency) to the decision executed by human drivers during merging tasks. Then, an entropy-based uncertainty quantification approach is developed to evaluate the changes in feature saliency across the decision-making process.

\subsection{LSTM-based Prediction}\label{subsec:LSTM}

\begin{figure}[t]
\centering
\includegraphics[width=\linewidth]{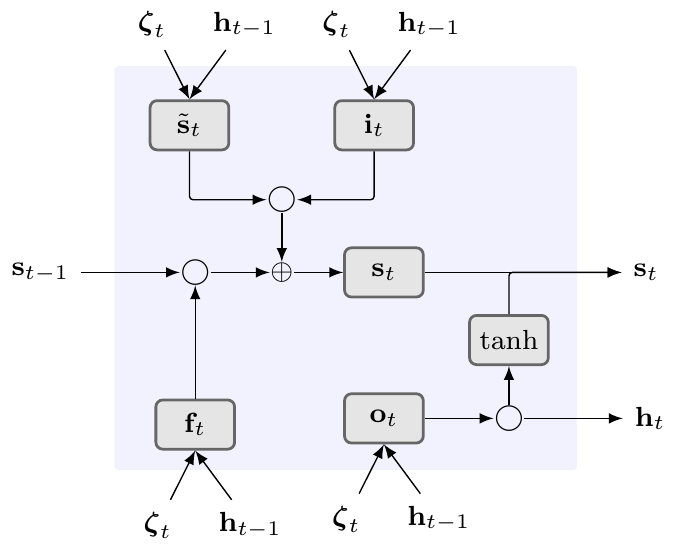}
\caption{The general structure of an LSTM memory cell with $\circ$ the Hadamard product.}
\label{fig:LSTM_Cell}
\end{figure}

We applied an LSTM-based model to predict the trajectory of the ego vehicle due to its powerful ability to integrate historical information into prediction. A standard LSTM network comprises an input layer, an output layer, and numerous hidden layers \cite{graves2013generating, yang2021fast}, as illustrated in Fig. \ref{fig:LSTM_Cell}. The LSTM memory cells are in the hidden layer. The forget gate $\mathbf{f}_t$ (determining information to remove), input gate $\mathbf{i}_t$ (specifying information to add), and output gate $\mathbf{o}_t$ (specifying information to output) are the three gates that each LSTM memory cell uses to retain its cell state $\mathbf{s}_t$. The execution procedures in an LSTM-based prediction can be completed in four steps. 

\begin{enumerate}
\item Firstly, the information that has to be removed from the previous cell state is determined by the activation values of the forget gate $\mathbf{f}_t$, calculated by

\begin{equation}\label{eq:LSTM_step1}
\mathbf{f}_t = \sigma (\mathbf{W}_{\boldsymbol{\zeta},\mathbf{f}}\boldsymbol{\zeta}_t + \mathbf{W}_{\mathbf{h},\mathbf{f}}\mathbf{h}_{t-1} + \mathbf{b}_\mathbf{f})
\end{equation}
where $\sigma(\cdot)$ is a sigmoid function with an output scale in $[0,1]$. We use a matrix $\mathbf{W}_{\ast,\dag}$ with subscripts to represent the weighted relationship between $\ast$ and $\dag$, and $\mathbf{b}_{\divideontimes}$ to represent the associated bias vector for gate $\divideontimes$. For example, $\mathbf{W}_{\boldsymbol{\zeta},\mathbf{f}}$ and $\mathbf{W}_{\mathbf{h},\mathbf{f}}$ are the weight matrices of the current driving data sample $\boldsymbol{\zeta}_t$ at time $t$ and the hidden cell state's output $\mathbf{h}_{t-1}$ at time $t-1$, respectively. $\mathbf{b}_f$ is the bias vector for boosting the model's adaptability to fitting data.

\item Secondly, the new cell state will determine how much information at the current time should be updated. The hyperbolic tangent function is used to compute candidate values $\tilde{\mathbf{s}}_t$ that may exist in the new cell state $\mathbf{s}_t$. What candidate values $\tilde{\mathbf{s}}_t$ should be added in the cell state $\mathbf{s}_t$ is then determined by the input gate's activation values $\mathbf{i}_t$, calculated as follows:

\begin{equation}\label{eq:LSTM_step2_S}
\tilde{\mathbf{s}}_t = \tanh(\mathbf{W}_{\boldsymbol{\zeta},\tilde{\mathbf{s}}}\boldsymbol{\zeta}_t + \mathbf{W}_{\mathbf{h},\tilde{\mathbf{s}}}\mathbf{h}_{t-1} + \mathbf{b}_{\tilde{\mathbf{s}}})
\end{equation}
and 
\begin{equation}\label{eq:LSTM_step2_i}
\mathbf{i}_t = \sigma (\mathbf{W}_{\boldsymbol{\zeta},\mathbf{i}}\boldsymbol{\zeta}_t + \mathbf{W}_{\mathbf{h},\mathbf{i}}\mathbf{h}_{t-1} + \mathbf{b}_\mathbf{i})
\end{equation}
where the settings of $\mathbf{W}_{\boldsymbol{\zeta},\tilde{\mathbf{s}}}, \mathbf{W}_{\mathbf{h},\tilde{\mathbf{s}}}, \mathbf{W}_{\boldsymbol{\zeta},\mathbf{i}}$, and $ \mathbf{W}_{\mathbf{h},\mathbf{i}} $ are the weight matrices defined in the same way as the ones in (\ref{eq:LSTM_step1}). $\mathbf{b}_{\tilde{\mathbf{s}}}$ and $\mathbf{b}_\mathbf{i}$ are corresponding bias terms.

\item Thirdly, the new cell states $\mathbf{s}_t$ are calculated by summing the previous and current information via

\begin{equation}\label{eq:LSTM_step3}
\mathbf{s}_t = \mathbf{f}_t \circ \mathbf{s}_{t-1} + \mathbf{i}_t \circ \tilde{\mathbf{s}}_t
\end{equation}
where $\circ$ denotes the Hadamard (elementwise) product. We elementwise multiply $\mathbf{f}_t$ with the previous cell state $\mathbf{s}_{t-1}$ to evaluate how much the previous information should be forgotten and multiply $\mathbf{i}_t$ and the candidate values $\tilde{\mathbf{s}}_t$ to determine how much the current information should be remembered. 

\item Lastly, the output $\mathbf{h}_t$ is derived according to (\ref{eq:LSTM_step4_h}) 
\begin{equation}\label{eq:LSTM_step4_h}
\mathbf{h}_t = \mathbf{o}_t \circ \tanh (\mathbf{s}_t)
\end{equation}
based on the updated activation values $\mathbf{o}_t$ using

\begin{equation}\label{eq:LSTM_step4_O}
\mathbf{o}_t = \sigma (\mathbf{W}_{\boldsymbol{\zeta}, \mathbf{o}}\boldsymbol{\zeta}_t + \mathbf{W}_{\mathbf{h},\mathbf{o}}\mathbf{h}_{t-1} + \mathbf{b}_{\mathbf{o}})
\end{equation}
where $\mathbf{W}_{\boldsymbol{\zeta}, \mathbf{o}}$ and $\mathbf{W}_{\mathbf{h},\mathbf{o}}$ are the matrices defined in the same way as the one in (\ref{eq:LSTM_step1}), and $\mathbf{b}_{\mathbf{o}}$ is the bias term.
\end{enumerate}

We process the sequential time-series driving data at each time $t$ to the LSTM network according to Equations (\ref{eq:LSTM_step1}) -- (\ref{eq:LSTM_step4_h}) under the architecture of the prediction network. This network structure consists of the first layer of $32$ LSTM memory cells, followed by two dense and time-distributed layers of $32$ and $16$ neurons, and a final dense output layer. We apply the Adam solver \cite{kingma2014adam} with a learning rate of $0.005$ and define the mean squared error (MSE) between the network prediction and the actual value as the training loss.

\subsection{Explanation of Predictions Using SHAP} \label{subsec:SHAP}
This section introduces a surrogate model-based explainable approach, SHapley Additive exPlanations (SHAP), to explain each feature's contribution to the deep learning-based model predictions. We denote the feature candidates as a feature set, $\mathcal{F} \equiv \boldsymbol{\zeta} = \{\zeta_{1}, \zeta_{2}, \dots, \zeta_{M} \} $ and the prediction as $\xi\in\mathbb{R}$. Here, $M=6$ represents the number of selected features as model inputs. Thus, the goal is to evaluate and explain the contribution value of each feature $\zeta_{i}$ to predictions $\xi$, 

\begin{equation}
\phi_{i}: (\zeta_{i}, \xi) \rightarrow \mathbb{R}.
\end{equation}
The predictive model in Section \ref{subsec:LSTM} with $\boldsymbol{\zeta}$ as inputs and $\xi$ as output is used to represent human drivers, thereby the obtained contribution value of each feature in the prediction model can interpret which feature is more salient for the driver to make decisions.

\subsubsection{Shapley Values}
Before introducing SHAP, we have to revisit the Shapley value briefly. One of the most classic ways in cooperative game theory to compute the contribution value is the Shapley regression method, which quantifies the feature importance in the model \cite{lipovetsky2001analysis}. The basic idea of this method is to retain the model on \textit{all} feature subset $\mathcal{S}\subseteq \mathcal{F}$. The Shapley value ($\phi_{i}\in\mathbb{R}$) is a real value and is utilized as feature attributions with a weighted average of all possible differences via

\begin{equation}\label{eq:shapley}
\phi_{i} = \sum_{\mathcal{S}\subseteq \mathcal{F}\setminus\{\zeta_{i}\}} \gamma\left[f(\boldsymbol{\zeta}_{\mathcal{S}\cup\{\zeta_{i}\}}) - f(\boldsymbol{\zeta}_{\mathcal{S}})\right]
\end{equation}
with the weight factor $\gamma$ counting the number of permutations of feature subset $\mathcal{S}$

\begin{equation*}
\gamma = \frac{|\mathcal{S}|!\left(|\mathcal{F}|-|\mathcal{S}|-1\right)!}{|\mathcal{F}|!}
\end{equation*}
where $|\mathcal{F}|=M$, $\boldsymbol{\zeta}_{\mathcal{S}}$ is the input features in the set $\mathcal{S}$, $f(\boldsymbol{\zeta}_{\mathcal{S}})$ and $f(\boldsymbol{\zeta}_{\mathcal{S}\cup\{\zeta_{i}\}})$ are the models trained with feature $\zeta_{i}$ absent and present, respectively. To calculate the contributions of features $\zeta_{i}$, we are considering the differences between these two models, reflecting the effect of adding $\zeta_{i}$ to a subset $\mathcal{S}$ on model prediction, which is represented by $f(\boldsymbol{\zeta}_{\mathcal{S}\cup\{\zeta_{i}\}}) - f(\boldsymbol{\zeta}_{\mathcal{S}})$. Then, we take a weighted average of this quantity across all possible subsets $\mathcal{S}\subseteq \mathcal{F}$, formulated by (\ref{eq:shapley}).

\subsubsection{SHAP}
In applications, SHAP provides a global explanation and local explainability for the causes of individual predictions by treating the Shapley value of a sample $\zeta_{i}$ as an additive feature importance score for prediction, which has three properties: local accuracy, missingness, and consistency \cite{madhikermi2019explainable}.

\begin{itemize}
\item \textbf{Local accuracy.} An explanation model (denoted as $g$) is a local surrogate model of the original model (denoted as $f$) and needs to match the output of the original model for the simplified inputs $\boldsymbol{\zeta}^{\prime}$, i.e., $f(\boldsymbol{\zeta}) = g(\boldsymbol{\zeta}^{\prime}) $ with

\begin{equation}\label{eq:shap_fx}
\begin{aligned}
g(\boldsymbol{\zeta}^{\prime}) & = \phi_0 + \sum_{i=1}^{M} \phi_{i} \zeta^{\prime}_{i} \\
     &= \mathrm{bias} + \sum \mathrm{contribution \ of \ each \ feature}
\end{aligned}
\end{equation}
where $\boldsymbol{\zeta}^{\prime}$ is related to its original feature $\boldsymbol{\zeta}$ by the mapping function $h_\zeta: \boldsymbol{\zeta} \rightarrow \boldsymbol{\zeta}^{\prime}$, and $\phi_{i}$ represents the attribution for a feature $\zeta_{i}$ in which the predicted value is increased (or decreased) by the positive (or negative) Shapely value, interpreting the original model. Specially, $\phi_0 = f(h_{\zeta}(0))$ denotes all simplified features are toggled off. The output of the original model $f$ can be approximated by adding up the impacts of all feature contributions.

\item \textbf{Missingness.} An absent feature is assigned a zero attribution according to the missingness property

\begin{equation}\label{eq:shap_missingness}
\zeta^{\prime}_{i} = 0 \Rightarrow \phi_{i} = 0
\end{equation}
where $\zeta^{\prime}_{i} \in \{0,1 \}^M$. More specifically, $\zeta^{\prime}_{i} =1$  when the feature input $\zeta^{\prime}_{i}$ is present or $\zeta^{\prime}_{i} =0$ when excluded. Since a missing feature multiplies $\zeta^{\prime}_{i} = 0$ will obtain a Shapley value of zero and not influence the local accuracy.

\item \textbf{Consistency.} The consistency property states that the marginal contribution of a feature value for the simplified inputs $\boldsymbol{\zeta}^{\prime}$ and the Shapley value will increase or stay the same with model change regardless of the other inputs \cite{lundberg2017unified}. 
\end{itemize}

\begin{figure}[t]
\centering
\includegraphics[width=\linewidth]{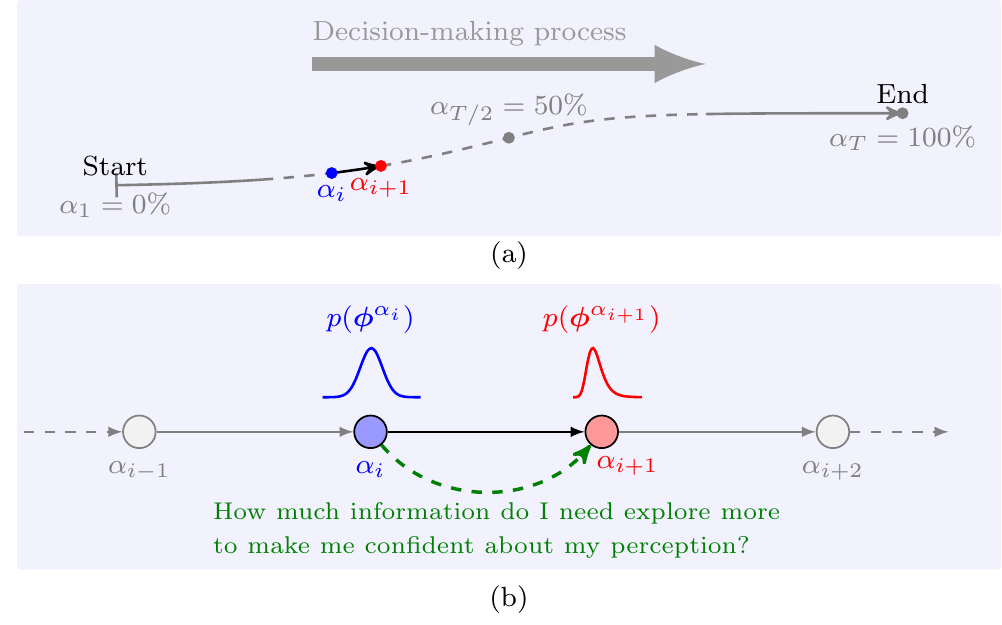}
\caption{Schematic diagram of (a) the evenly partitioned decision-making process using percentages $\{\alpha_{i}\}$ and (b) the distribution of feature saliency at adjacent merge percentages $\alpha_{i}, \alpha_{i+1}\in[0, 100]\%$ and the entropy-based uncertainty quantification.}
\label{fig:entropycomputation}
\end{figure}

In summary, the three above properties ensure that the Shapley value $\phi_{i}$ can finally be computed using (\ref{eq:shapley}). The SHAP framework achieves reliability by comprehensively calculating a specific feature's contribution to a model prediction from all possible combinations, quantifying the influence as positive or negative. Most importantly, SHAP is consistent in the importance of a particular feature even when the model changes. Therefore, we quantify the saliency of driving features listed in Table \ref{tab:definition} at three key moments (i.e., $\boldsymbol{\phi}^{\alpha=0\%}$, $\boldsymbol{\phi}^{\alpha=50\%}$, and $\boldsymbol{\phi}^{\alpha=100\%}$) to explain the predictions of the LSTM-based model. $\boldsymbol{\phi}^{\alpha} \in\mathbb{R}^{M}$ is a vectorized SHAP value at the key moment, $\alpha$, and represents the estimated feature saliency. The schematic diagram of Fig. \ref{fig:entropycomputation}(a) illustrates the discretized moments in the decision-making process during merging tasks.

\subsubsection{Distribution of SHAP Values} 
\begin{figure}[t]
\centering
\includegraphics[width=\linewidth]{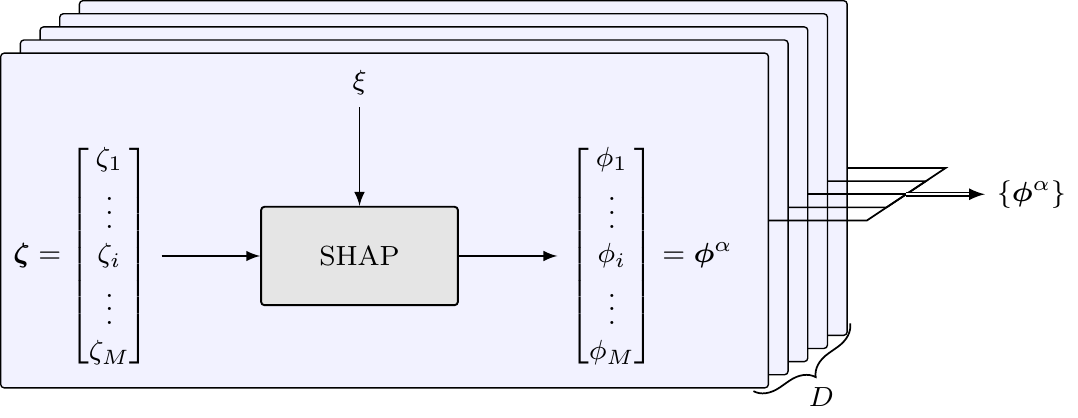}
\caption{Schematic diagram of the SHAP values set $\{\boldsymbol{\phi}^{\alpha}\}$ for all demonstrations ($D=796$) at a specific decision-making moment $\alpha$. Each plate represents a merge demonstration.}
\label{fig:SHAP}
\end{figure}

For each merge demonstration, we use the developed SHAP tool to estimate the feature saliency ($\boldsymbol{\phi}$) at a fixed decision-making moment $\alpha$ to their model predictions, as illustrated in Fig. \ref{fig:SHAP}. However, the variant duration of merge demonstrations prevents comparison among them. To comprehensively analyze all merge events, we use percentage values to describe each demonstration's decision-making process, allowing aligning all demonstrations, as illustrated in Fig. \ref{fig:entropycomputation}(a). The SHAP values of all demonstrations at a specific percentage point $\alpha$ form a set $\{\boldsymbol{\phi}^{\alpha}\}$ characterized by a distribution $p(\boldsymbol{\phi}^{{\alpha}})$, where $\alpha\in[0, 100] \%$ represents how much in percentage the human driver has completed the decision-making process. For instance, $p(\boldsymbol{\phi}^{{50\%}})$ represents the statistical information of feature saliency at half of the decision-making process for all demonstrations.

\subsection{Entropy-based Uncertainty Quantification} \label{subsec:Entropy Evaluation}

The above section provides a distribution of feature saliency $p(\boldsymbol{\phi}^{{\alpha}})$, which probabilistically represents what kind of information (i.e., feature candidates) human drivers use to make efficient decisions at specific moments $\alpha$. This section will focus on how the information would evolve over the decision-making process during the merge task, as shown in Fig. \ref{fig:entropycomputation}(b). In other words, we are interested in how human drivers' beliefs about the environment change based on their perceptual information to make a trustworthy decision. The following facts inspire this assertion: A small change in information exploitation indicates that the human driver would be more confident in their current perceptions about the environment since they only need to seek less or no additional information (i.e., slightly change their beliefs) in order to make proper decisions. On the contrary, a significant change in information exploitation indicates that human driver has less confidence in their current perceptions. Therefore, they need to seek more informative cues to update and increase their beliefs and make decisions. Therefore, \textit{by checking how much information used for decision-making is changed, we can indirectly infer how the perceptual uncertainty in the human driver's mind varies over the decision-making process}. In this paper, to quantify the changes in the information exploited to make decisions, we applied two types of entropy-based approaches: one is the Kullback-Leibler (KL) divergence \cite{kullback1951information, csiszar1975divergence, lu2022instance}, and the other one is mutual information \cite{cover1999elements, kraskov2004estimating, altun2017road}.

\begin{figure}[t]
\centering
\includegraphics[width=\linewidth]{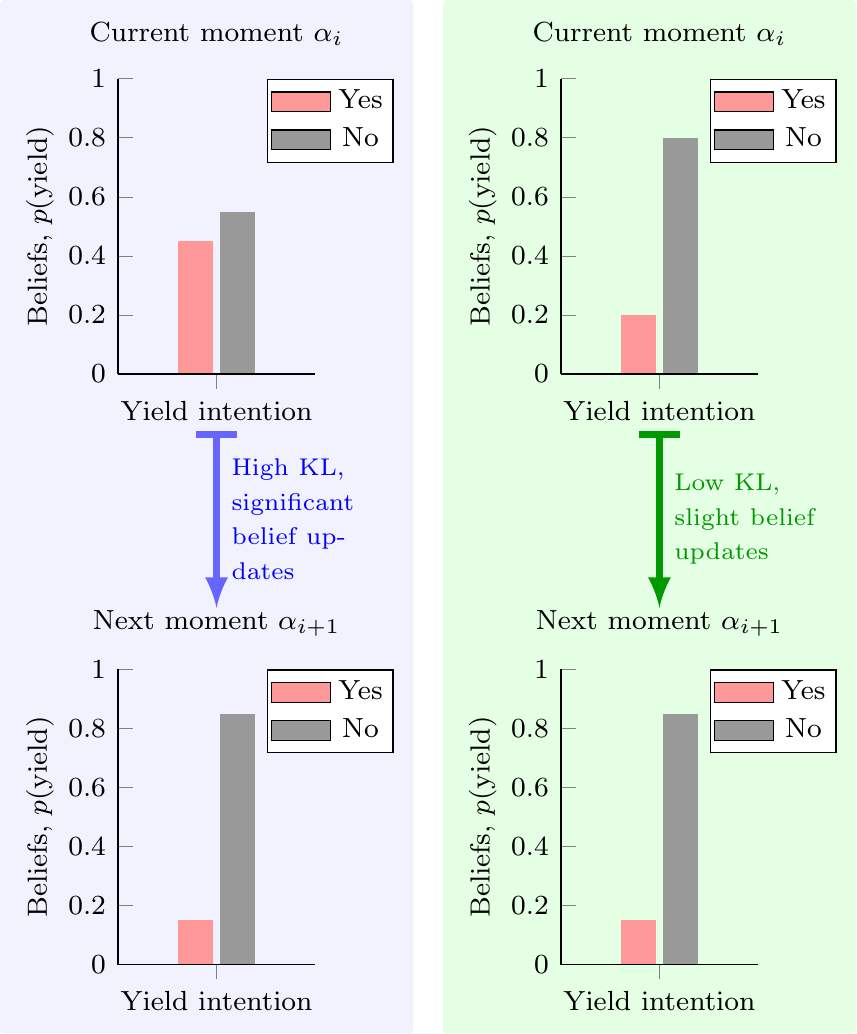}
\caption{Illustration of changes in the ego driver's internal beliefs about the purpose of other vehicles from the current to the next merge moment. Two different beliefs in the current merging moment are compared.}
\label{eq:KLexplanation}
\end{figure}

\subsubsection{KL Divergence}
We introduced the KL divergence to measure the changes in the distribution of feature saliency across the decision-making process, starting at $\alpha=0\%$ and ending with $\alpha=100\%$. The KL divergence is a non-negative relative entropy and is used to measure the difference between $p(\boldsymbol{\phi}^{\alpha_{i}})$ and $p(\boldsymbol{\phi}^{\alpha_{i+1}})$ at two adjacent merge moments $\alpha_{i}$ and $\alpha_{i+1}$. In this paper, the KL divergence in a discrete case is computed by

\begin{equation}\label{eq:KL_d}
D_{\mathrm{KL}}\left(p(\boldsymbol{\phi}^{\alpha_{i}}) \parallel p(\boldsymbol{\phi}^{\alpha_{i+1}})\right) = \sum_{\boldsymbol{\phi}} p(\boldsymbol{\phi}^{\alpha_{i}}) \log \frac{p(\boldsymbol{\phi}^{\alpha_{i}})}{p(\boldsymbol{\phi}^{\alpha_{i+1}})}
\end{equation}

From the perspective of Bayesian inference, the KL divergence (\ref{eq:KL_d}) measures the information loss when the condition changes from the prior probability distribution $p(\boldsymbol{\phi}^{\alpha_{i}})$ to the posterior probability distribution $p(\boldsymbol{\phi}^{\alpha_{i+1}})$ \cite{wolpert1995estimating}. Thus, the intuition behind the KL divergence in this work can be interpreted using Fig. \ref{eq:KLexplanation}. A slight KL divergence indicates that the human driver does not need a significant update of their beliefs\footnote{Here, updating beliefs can be viewed as a Bayesian inference procedure; that is, inferring/updating the posterior distribution once observed new samples, given a prior distribution.} of the environment based on their current perception to make next-step decisions; that is, the human driver is highly confident in their perception of the environment, corresponding to a low perceptual uncertainty. The green line in Fig. \ref{eq:KLexplanation} can interpret this. Conversely, a large KL divergence indicates that the human driver is highly uncertain about the environment (i.e., not trustworthy about their perception) and needs to seek more valuable information to reduce the perceptual uncertainty, thus resulting in a significant KL divergence, corresponding to the blue line in Fig. \ref{eq:KLexplanation}.

\subsubsection{Mutual Information} 
In addition to focusing on the changes in feature saliency using KL divergence, we are also interested in the \textit{dependence} of feature saliency between two adjacent moments, $\boldsymbol{\phi}^{\alpha_{i}}$ and $\boldsymbol{\phi}^{\alpha_{i+1}}$. The dependency is referred to as the amount of information needed at the present moment to make a prediction and decision for the next adjacent moment, quantified through the joint distribution relative to the marginal distribution based on the mutual information theory \cite{cover1999elements},

\begin{equation}\label{eq:Mutual_information}
\begin{split}
I(\boldsymbol{\phi}^{\alpha_{i}};\boldsymbol{\phi}^{\alpha_{i+1}})
&= H(\boldsymbol{\phi}^{\alpha_{i}},\boldsymbol{\phi}^{\alpha_{i+1}}) - H(\boldsymbol{\phi}^{\alpha_{i}}|\boldsymbol{\phi}^{\alpha_{i+1}}) \\
 & \ \ \ - H(\boldsymbol{\phi}^{\alpha_{i+1}}|\boldsymbol{\phi}^{\alpha_{i}})
\end{split}
\end{equation}
with
\begin{equation*}\label{eq:entropy}
H(\boldsymbol{\phi}^{\alpha}) = - \sum_{\boldsymbol{\phi}} p(\boldsymbol{\phi}^{\alpha}) \log p(\boldsymbol{\phi}^{\alpha})
\end{equation*}
where $H(\cdot)$ are the marginal entropies, $H(\cdot|\cdot)$ are the conditional entropies, and $H(\cdot,\cdot)$ is the joint entropy. Equation (\ref{eq:Mutual_information}) indicates that the mutual information value reflects the perceptual uncertainty reduction at a specific moment by knowing features at other moments. More specifically, independence between $\boldsymbol{\phi}^{\alpha_{i}}$ and $\boldsymbol{\phi}^{\alpha_{i}}$ indicates that knowing $\boldsymbol{\phi}^{\alpha_{i}}$ does not provide any information about $\boldsymbol{\phi}^{\alpha_{i+1}}$ and vice versa. On the other hand, if both $\boldsymbol{\phi}^{\alpha_{i}}$ and $\boldsymbol{\phi}^{\alpha_{i+1}}$ are deterministic functions of each other, then they share all information --- knowing $\boldsymbol{\phi}^{\alpha_{i}}$ determines the value of $\boldsymbol{\phi}^{\alpha_{i+1}}$ and vice versa. In this case, the mutual information refers to the perceptual uncertainty contained in feature saliency $\boldsymbol{\phi}^{\alpha_{i}}$ and $\boldsymbol{\phi}^{\alpha_{i+1}}$, which is bidirectional and symmetrical.

\section{Result Analysis and Discussion}
This section first assesses the prediction performance of our developed LSTM-based model and then explains the prediction results quantitatively and qualitatively using SHAP with various evaluation metrics. A systematic analysis of the uncertainty changes is then provided.

\subsection{Time-Series Prediction Performance}
\subsubsection{Performance Metrics}
The prediction capability of a model is essential for a reliable explanation of salient features' influences on this model. We quantified the prediction performance of our developed LSTM model using the root-mean-square error (RMSE, $\epsilon_{\mathrm{RMSE}}$) based on each demonstration and the evaluation score of MSE, $\beta_{\mathrm{MSE}}$. 

\paragraph{RMSE} For individual merge demonstrations, RMSE describes the unbiased estimation of the error variance (accuracy) and is computed by

\begin{equation}\label{eq:RMSE}
\begin{split}
\epsilon_{\mathrm{MSE}} &= \frac{1}{N_{d}}\sum_{t=1}^{N_{d}} \left(\hat{\xi}_t - \xi_t\right)^2 \\
\epsilon_{\mathrm{RMSE}} &= \sqrt{\epsilon_{\mathrm{MSE}}}
\end{split}
\end{equation}
where $N_{d}$ represents the total number of data samples in demonstration $d$, where $d=1,2,\dots, D$, $\hat{\xi}_t$ is the prediction at time $t$ according to (\ref{eq:LSTM_step4_h}), and $\xi_t$ is the ground truth.

\paragraph{Evaluation Score of MSE} We also introduce the evaluation score of MSE \cite{mccuen2006evaluation} to quantify the prediction stability. For each demonstration $d$, it is computed by

\begin{equation}\label{eq:s_MSE}
\beta_{\mathrm{MSE}}=\frac{\epsilon_{\mathrm{MSE}} - \epsilon_{\mathrm{MSE}}^{\mathrm{ref}}}{0 - \epsilon_{\mathrm{MSE}}^{\mathrm{ref}}}
\end{equation}
with 
\begin{equation}
\epsilon_{\mathrm{MSE}}^{\mathrm{ref}}=\frac{1}{N_{d}} \sum_{t=1}^{N_{d}} \left(\bar{\xi} - \xi_t\right)^2
\end{equation}
where $\bar{\xi}$ is the mean value of $\xi_t$ for $t\in(1,N_{d})$. We use the average MSE ($\bar{\beta}_{\mathrm{MSE}}$) and average evaluation score ($\bar{\epsilon}_{\mathrm{RMSE}}$) of all merge demonstrations to evaluate the performance, as listed in Tables \ref{table:score_approach} and \ref{table:model_generalization}. The average metrics can embody the model performance: a high positive $\bar{\beta}_{\mathrm{MSE}}$ and low $\bar{\epsilon}_{\mathrm{RMSE}}$ implies a good prediction performance of our model. We also compare the prediction results with other data-driven models developed based on Gaussian mixture regression (GMR)/Gaussian mixture models (GMM), hidden Markov models (HMM), and their synthesizing \cite{wang2021uncovering}, as shown in Table \ref{table:score_approach}.

\renewcommand\arraystretch{1.5} 
\begin{table}[t]
	\centering
	\caption{Performance of Different Approaches with Same Input Features ($\Delta v_{x}^{\mathrm{lead}}$, $\Delta {x}^{\mathrm{lag}}$, $v_x^{\mathrm{ego}}$) and Output Feature $v_y^{\mathrm{ego}}$}\label{table:score_approach}

	\begin{tabular}{ c | c c c}

		\hline \hline
		Approach & GMM-GMR & HMM-GMR & Ours\\
		
		\hline
		 $\bar{\beta}_{\mathrm{MSE}}$ & 0.485 & 0.686& \textbf{0.890} ($\uparrow$29.7\%)\\
		
		 $\bar{\epsilon}_{\mathrm{RMSE}}$ & 0.065 & 0.059 & \textbf{0.050} ($\downarrow$15.3\%) \\
		\hline \hline
	\end{tabular}
\end{table}

\newcommand{\tabincell}[2]{\begin{tabular}{@{}#1@{}}#2\end{tabular}}

\renewcommand\arraystretch{1.5} 
\begin{table}[t]
	\centering
	\caption{Our Prediction Model Performance with Different Input and Output Features}\label{table:model_generalization}

	\begin{tabular}{ c| c  c }

		\hline \hline
		Inputs & Output & $\bar{\beta}_{\mathrm{MSE}}$ \\
		
		\hline
		
		 $\Delta v_{x}^{\mathrm{lead}}, \Delta {x}^{\mathrm{lag}},v_x^{\mathrm{ego}}$ & $v_y^{\mathrm{ego}}$ & 0.890 \\
		
		\hline
		
		 \tabincell{c}{$\Delta {x}^{\mathrm{lead}}$, $\Delta v_{x}^{\mathrm{lead}}$, $v_x^{\mathrm{ego}}$, $v_y^{\mathrm{ego}}$, $\Delta {x}^{\mathrm{lag}}$, $\Delta v_x^{\mathrm{lag}}$} & $\Delta {x}^{\mathrm{end}}$ & 0.995 \\
		
		\hline \hline
	\end{tabular}
\end{table}

\subsubsection{Results}
The comparative performance in Table \ref{table:score_approach} indicates that our developed LSTM-based prediction model has superior performance in terms of $\bar{\beta}_{\mathrm{MSE}}$ and $\bar{\epsilon}_{\mathrm{RMSE}}$. More specifically, compared with HMM-GMR and GMM-GMR, our model's prediction stability and accuracy are improved by $29.7\%$ and $15.3 \%$, respectively. Although HMM-GMR is specifically interpretable for its model parameters, its prediction performance is lower than our deep learning-based model.
In addition, we also evaluate our model with different input and output features to verify its generalization performance. Different outputs make it unfair to compare prediction errors with different units. Therefore, a fair comparison needs a dimensionless metric such as $\bar{\beta}_{\mathrm{MSE}}$. Experiment results in Table \ref{table:model_generalization} show that our model can still obtain a relatively stable prediction even with different features for inputs and outputs. 

\begin{figure}[t]
\centering
\includegraphics[width = \linewidth]{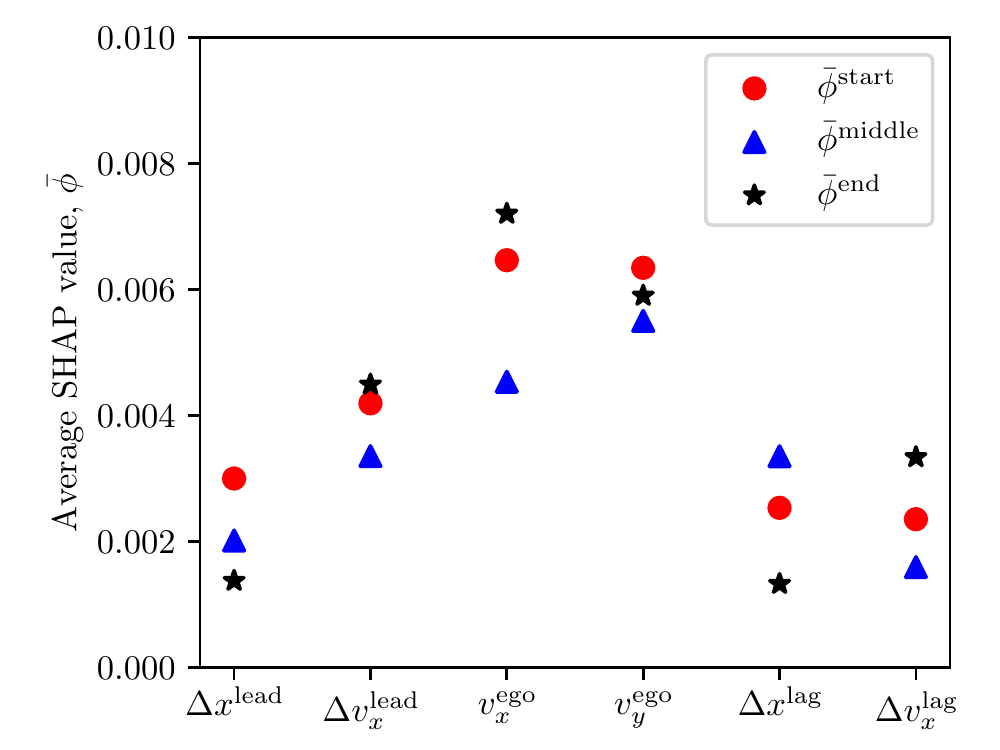}
\caption{The average SHAP values ($\bar{\phi}$) of all features at the three key merge moments (start, middle, and end).}
\label{fig:meanshap}
\end{figure}

\subsection{Changes in Feature Saliency for Decision-Making} \label{subsec:feature_saliency}

\begin{figure}[t]
\centering
\subfloat[Start moment, $\alpha = 0\%$]{\label{level2.sub.1}
\includegraphics[width=0.98\linewidth]{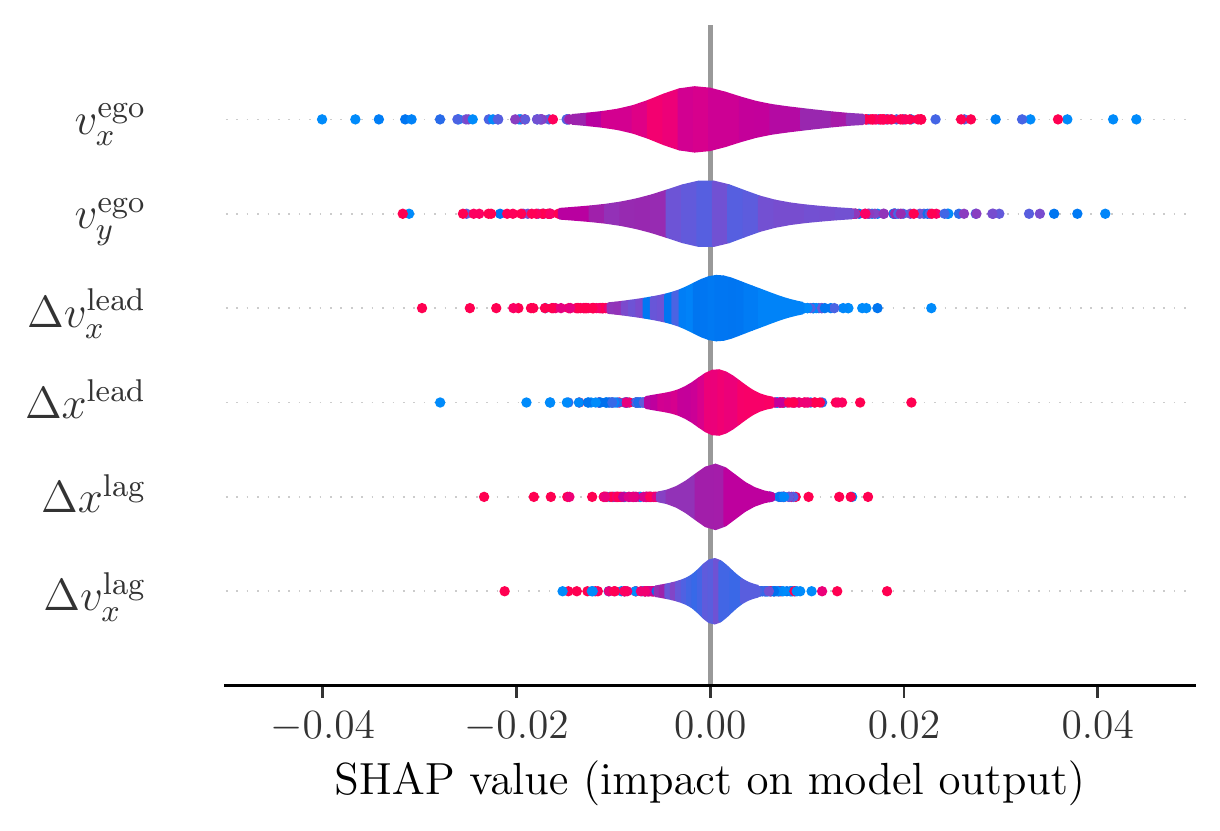}}\\
\subfloat[Middle moment, $\alpha = 50\%$]{\label{level2.sub.2}
\includegraphics[width=0.98\linewidth]{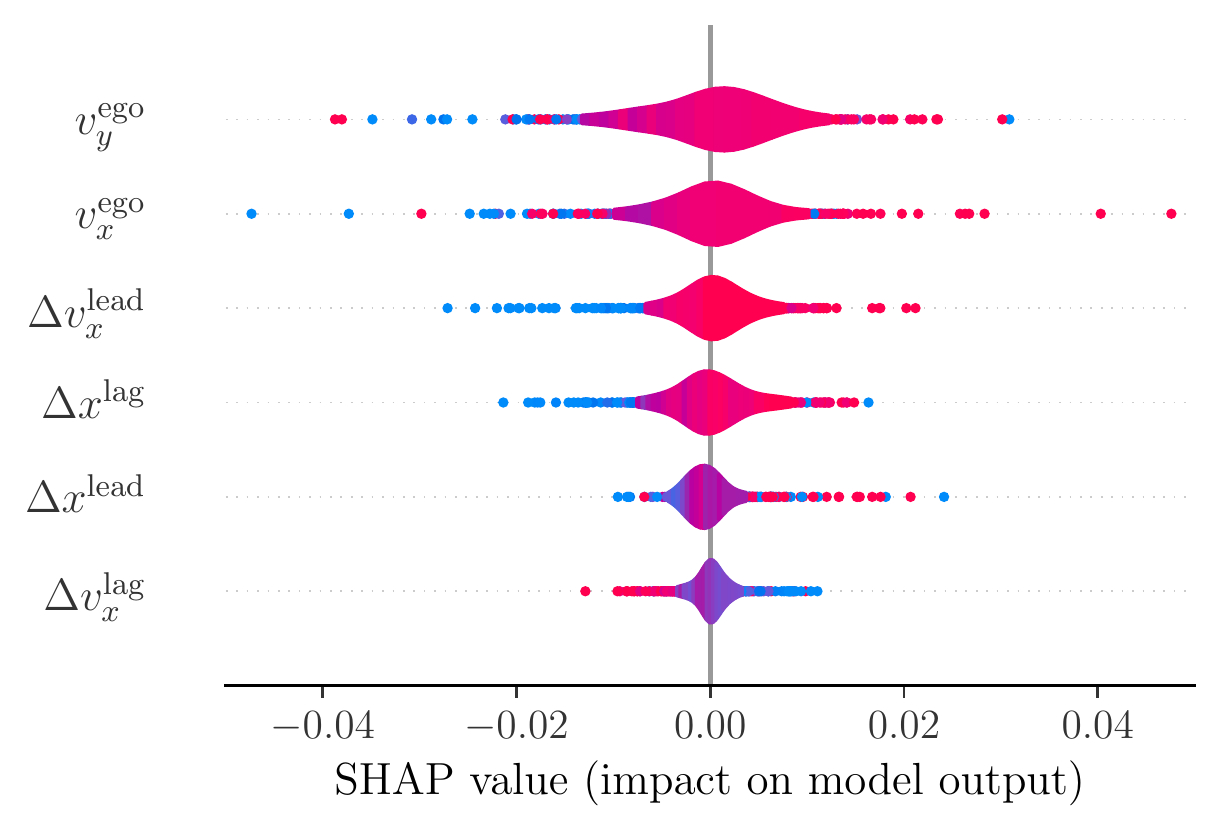}} \\
\subfloat[End moment, $\alpha = 100\%$]{\label{level2.sub.3}
\includegraphics[width=0.98\linewidth]{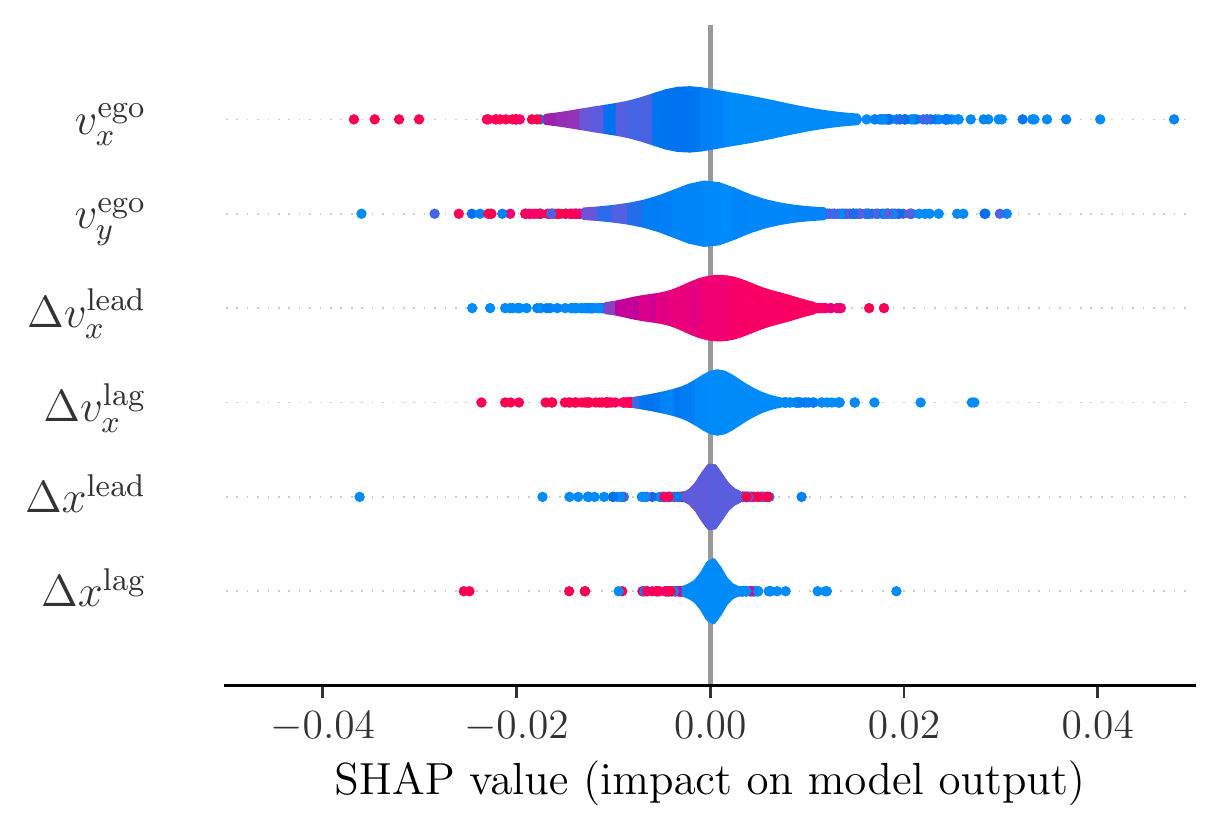}} \\
\caption{The violin plot of Shapley values at different key moments with colors representing feature values. Red (blue) represents the associated features' high (low) value.}
\label{fig:violin_shap}
\end{figure}

We quantitatively evaluate the SHAP's capability to explain predictions across the three key decision-making moments to compare with the results in \cite{wang2022on}. It should be mentioned that although we only choose three key decision-making moments for analysis (Fig. \ref{fig:meanshap}), the SHAP approach can predict and explain the results in a continuous time domain. Therefore, we set the input and output features of the prediction model consistent with those in \cite{wang2022on}.  We visualize the feature saliency at each key decision-making moment for convenient analysis using \texttt{Violin} plots, as shown in Fig. \ref{fig:violin_shap}. Red (blue) represents the corresponding features' high (low) value. The value of the horizontal axis represents the impact on model output $\Delta x^{\mathrm{end}}$, and positive (negative) values would make output larger (smaller).  In what follows, we will discuss and analyze the experiment results at the key decision-making moments.

\subsubsection{Start Moment, $\alpha=0\%$} 
Fig. \ref{fig:meanshap} indicates that the top-two salient features at the start moment are $v_x^{\mathrm{ego}}$ and $v_y^{\mathrm{ego}}$, with the highest values of $\bar{\phi}$, implying that human drivers make decisions mainly based on the ego vehicle's longitude and lateral speeds when starting to merge. The difference between them is that a large value of $v_{x}^{\mathrm{ego}}$ would shorten the distance to the end of the ramp since the corresponding SHAP values are red and negative. In addition, most SHAP values of $\Delta {x}^{\mathrm{lead}}$ are larger than zero, implying that its influence on model outputs is positive but only when the feature value is high since the color is red. 

\subsubsection{Middle Moment, $\alpha=50\%$}
Fig. \ref{fig:meanshap} indicates that human drivers mainly depend on the top-three features, $v_x^{\mathrm{ego}}$, $v_y^{\mathrm{ego}}$, and $\Delta v_{x}^{\mathrm{lead}}$, to make decisions in the middle moment. The red colors in Fig. \ref{fig:violin_shap}(b) imply that all these three salient features have high feature values. Conversely, the human driver would pay more attention to the distance difference with the lag vehicle ($\Delta {x}^{\mathrm{lag}}$) while pay less attention to the speed difference ($\Delta v_{x}^{\mathrm{lag}}$) with the lag vehicle. This behavioral decision is reasonable and consistent with the actual driving knowledge of human drivers in real congested traffic since a fast and decisive action can deliver clear and informative messages to surrounding vehicles and allow other vehicles to react, such as yielding, thus creating a trustworthy interaction.

\subsubsection{End Moment, $\alpha=100\%$}
Fig. \ref{fig:meanshap} shows that human drivers in the ego vehicle would pay more attention to their speed and the difference from the lead vehicle when the decision to merge ends since $v_x^{\mathrm{ego}}$, $v_y^{\mathrm{ego}}$, and $\Delta v_{x}^{\mathrm{lead}}$ reach high average SHAP values. Blue colors in Fig. \ref{fig:violin_shap}(c) imply that low feature values of $v_x^{\mathrm{ego}}$ and $v_y^{\mathrm{ego}}$ impact the distance to the endpoint of the merge decision but negatively impact the model output since most average SHAP values are negative. 
All these results on feature saliency are consistent with the conclusions in \cite{wang2022on}.

In summary, the developed deep learning-based prediction model and the explainable AI approach (i.e., SHAP) can obtain satisfying performance regarding prediction accuracy and model explanation. The verification above lays a solid foundation for analyzing the perceptual uncertainty based on the model explanation in continuous space. 

\begin{figure}[t]
\centering
\includegraphics[width=\linewidth]{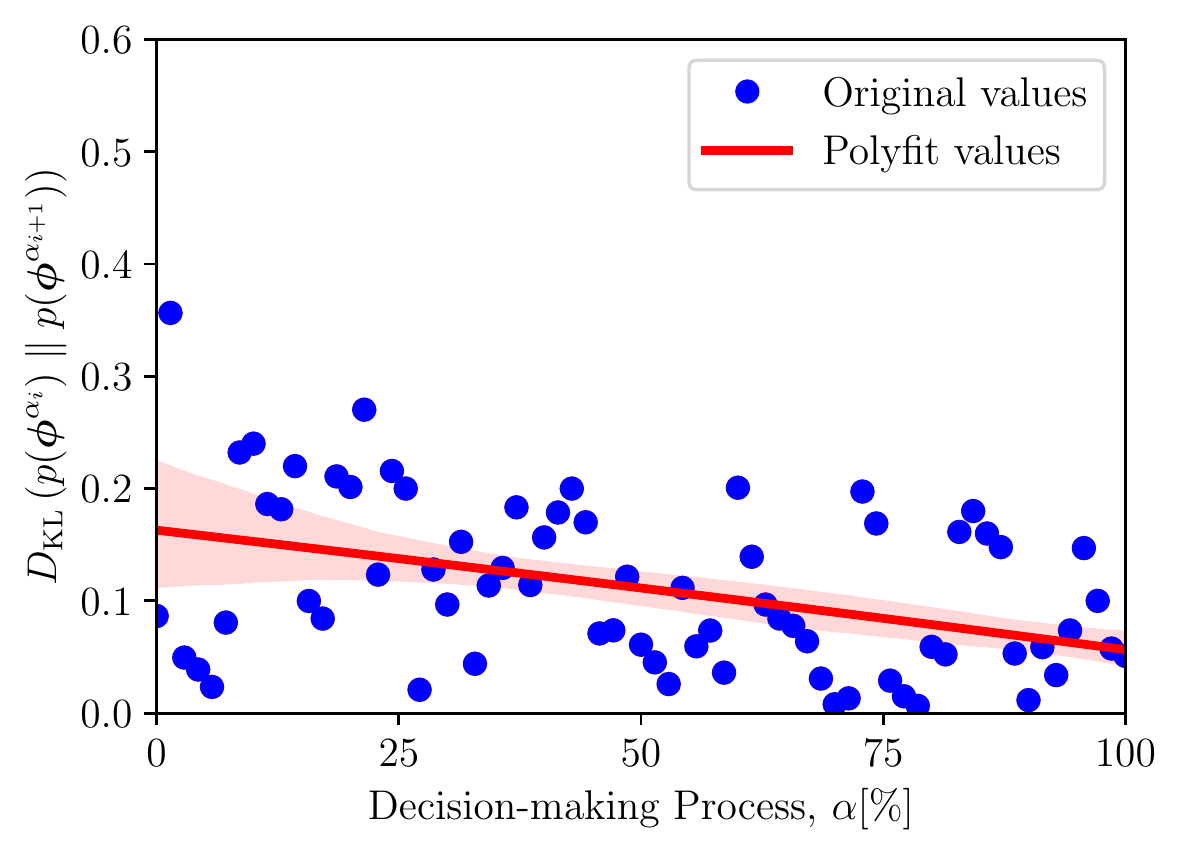}
\caption{The values and trends of KL divergence $D_{\mathrm{KL}}\left(\boldsymbol{\phi}^{\alpha_{i}}||\boldsymbol{\phi}^{\alpha_{i+1}}\right)$ over the entire decision-making process of merging tasks.}
\label{fig:KL_Divergence}
\end{figure}

\subsection{Uncertainty Change Analysis with Entropy}

\subsubsection{KL Divergence}
Section \ref{subsec:Entropy Evaluation} introduces the approaches to analyzing the changes in human drivers' perceptual uncertainty during decision-making based on the distribution of SHAP values of salient features. Fig. \ref{fig:KL_Divergence} illustrates the values of KL divergence $D_{\mathrm{KL}}(\boldsymbol{\phi}^{\alpha_{i}}||\boldsymbol{\phi}^{\alpha_{i+1}})$ across the decision-making moments $\alpha$, the scattered blue dots are the original KL values, and the solid red line is the fitting result. It demonstrates that the uncertainty between adjacent moments gradually decreases as the human driver continues to merge. 

The closer they are to the end of the decision-making procedure, the more confident they are in their perception. The human agent gradually tries to make their attention to features more compressed when interacting with the environment. Specifically, when starting to merge, the agent is highly uncertain of (i.e., low belief) other vehicles' intentions to yield or not. To reduce uncertainty, the human driver has to seek more information to update their beliefs about the environment by focusing on various environmental features, which causes a more significant KL divergence between the distribution of salient features at adjacent moments. With more information and updated beliefs about the environment as the merging process continues, the human driver can internally ensure that some potential features are more valuable to infer other vehicles' intentions and some are not. In such a way, it is only necessary for human drivers to focus on these salient features, resulting in a low uncertainty of the environment. The human driver's concentration on the surrounding objects increasingly narrows as they get closer to the end of decision-making, causing a steady order of salient features, i.e., a low uncertainty. In addition, the uncertainty variance is also decreasing over the decision-making process, as illustrated by the red highlighted region in Fig. \ref{fig:KL_Divergence}.

\begin{figure}[t]
\centering
\includegraphics[width=\linewidth]{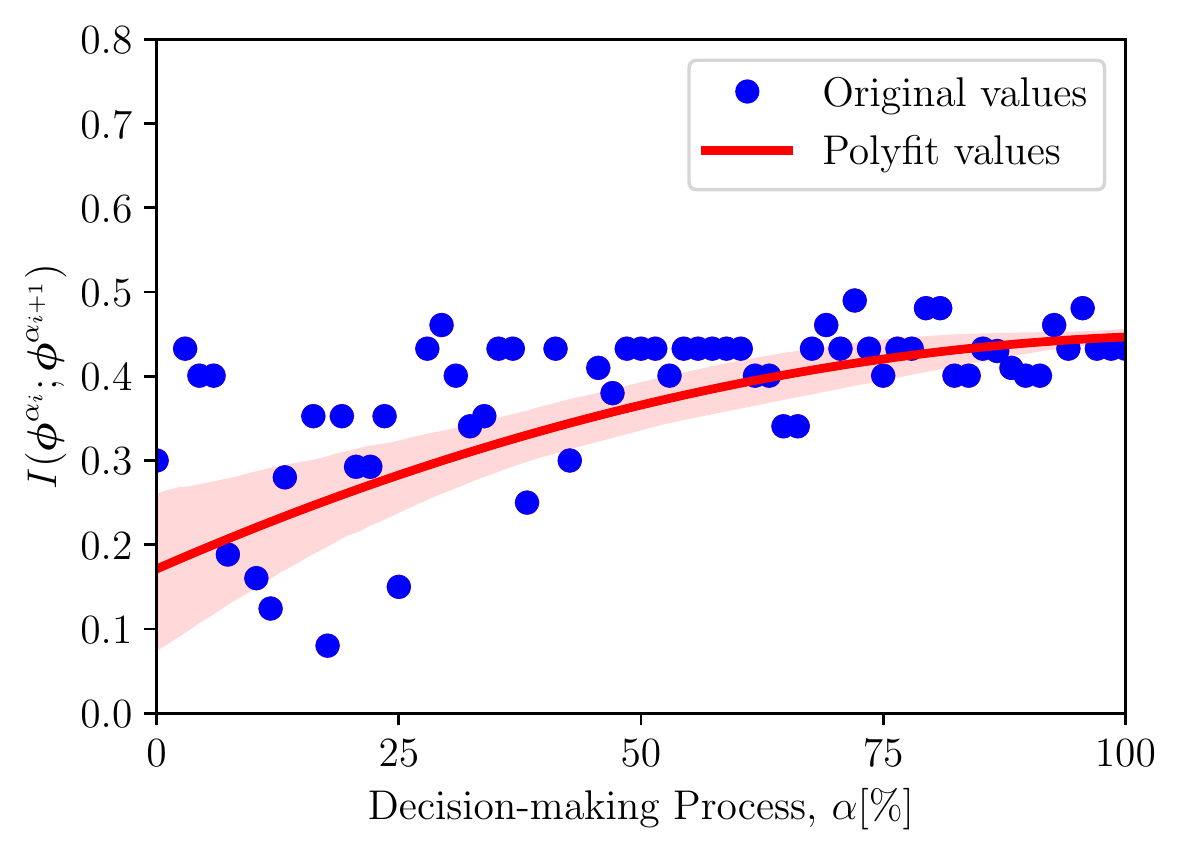}
\caption{The values and trends of mutual information $I\left(\boldsymbol{\phi}^{\alpha_{i}};\boldsymbol{\phi}^{\alpha_{i+1}}\right)$ over the entire decision-making process of merge tasks.}
\label{fig:Mutual_Information}
\end{figure}

\subsubsection{Mutual Information}
Fig. \ref{fig:Mutual_Information} illustrates the mutual information-based changes in the distribution of salient features across adjacent decision-making moments. The value (red line) of mutual information is low at the start of the merge decision with a significant variance (i.e., the red region is wide), indicating that the dependence of the salient features between adjacent moments is weak; that is, the ranked salient features have significant changes in their orders. As the merge behavior continues, there is a progressive increase in the dependence of salient features between adjacent decision-making moments $\alpha_{i}$ and $\alpha_{i+1}$. The intuition behind this trend is that the human driver progressively and selectively narrows the number of potential features to make decisions and treats two or three features as salient. In other words, the uncertainty about their perceptions or changes of salient features over the next moment continuously decreases conditioning on the environment at the last moment. An increase (decrease) in information-seeking behavior will accompany high (low) uncertainty. Reducing perceptual uncertainty in the interaction process will reduce the uncertainty of future behavior. This phenomenon is consistent with the above analysis of KL divergence.

\section{Conclusion}

This paper provided a general framework for explaining the human driver's sequential decisions in an interactive scenario from the perspective of perceptual uncertainty reduction by integrating explainable AI techniques (i.e., SHAP) and entropy-based uncertainty quantification. We empirically demonstrated that when merging into highways from the ramp, human drivers will seek more information to update and increase their beliefs and confidence about their perception of the environment (i.e., other vehicles' purpose) to reach a certain level on which they can make a trustworthy decision. During decision-making, human drivers selectively target several salient features and gradually narrow the number of selected features to make efficient and trustful decisions. In other words, the ego driver's perceptual uncertainty about the other agent's purpose would progressively decrease from the initial to the final stage. This conclusion is consistent with the free-energy principle \cite{friston2006free} and has the potential to comprehend the exploration-exploitation trade-off in reinforcement learning \cite{tschantz2020reinforcement}.





%




\section*{Acknowledgment}

We are grateful to all authors for their experiments and discussions. \textbf{Funding:} This work was supported by the National Key R\&D Program of China (2021YFB1716200) and an IVADO Postdoctoral Fellowship, Canada. \textbf{Author Contributions:} Conceptualization: W. Wang, H. Wang, and L. Sun; Methodology: H. Wang and W. Wang; Mathematical modeling: H. Wang and W. Wang; Experiments: H. Wang and H. Liu; Experimental data analysis: H. Wang and W. Wang; Supervision: H. Liu and L. Sun; Writing: H. Wang and W. Wang. \textbf{Competing interests:} The authors declare that they have no competing interests.

\ifCLASSOPTIONcaptionsoff
  \newpage
\fi




\bibliographystyle{IEEEtran.bst}
\bibliography{reference}

%

%


\begin{IEEEbiography}[{\includegraphics[width=1in,height=1.25in,clip,keepaspectratio]{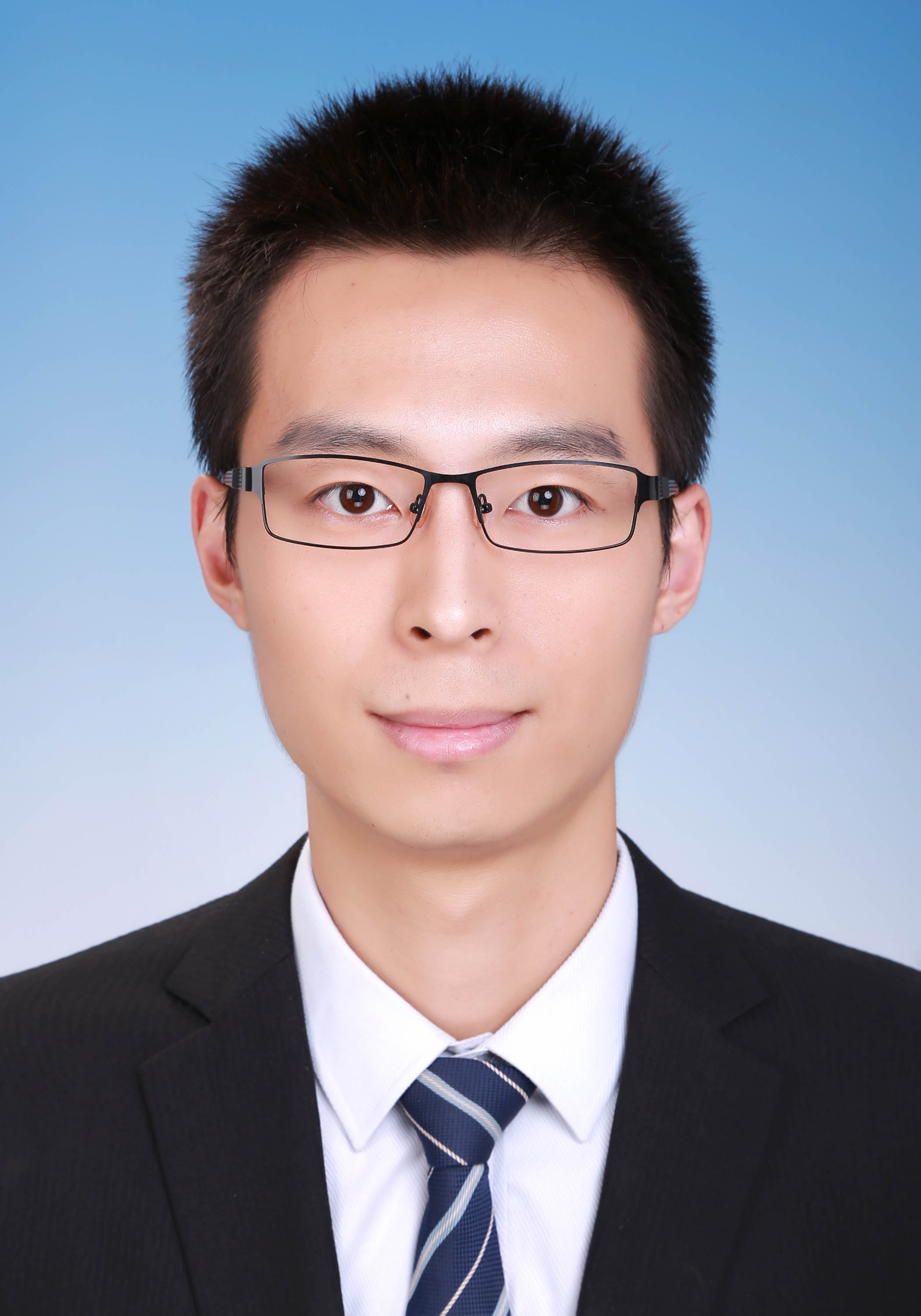}}]{Huanjie Wang}
received the Ph.D. degree from the School of Mechanical Engineering, Beijing Institute of Technology, China, in 2021. He was a research scholar at the University of California at Berkeley from 2018 to 2020. He is currently an assistant professor with the College of Intelligent Machinery, Faculty of Materials and Manufacturing, Beijing University of Technology, Beijing, China. His research interests include unmanned system platforms and robotics, situational awareness, driver behavior, decision-making, and machine learning.
\end{IEEEbiography}




\begin{IEEEbiography}[{\includegraphics[width=1in,height=1.25in,clip,keepaspectratio]{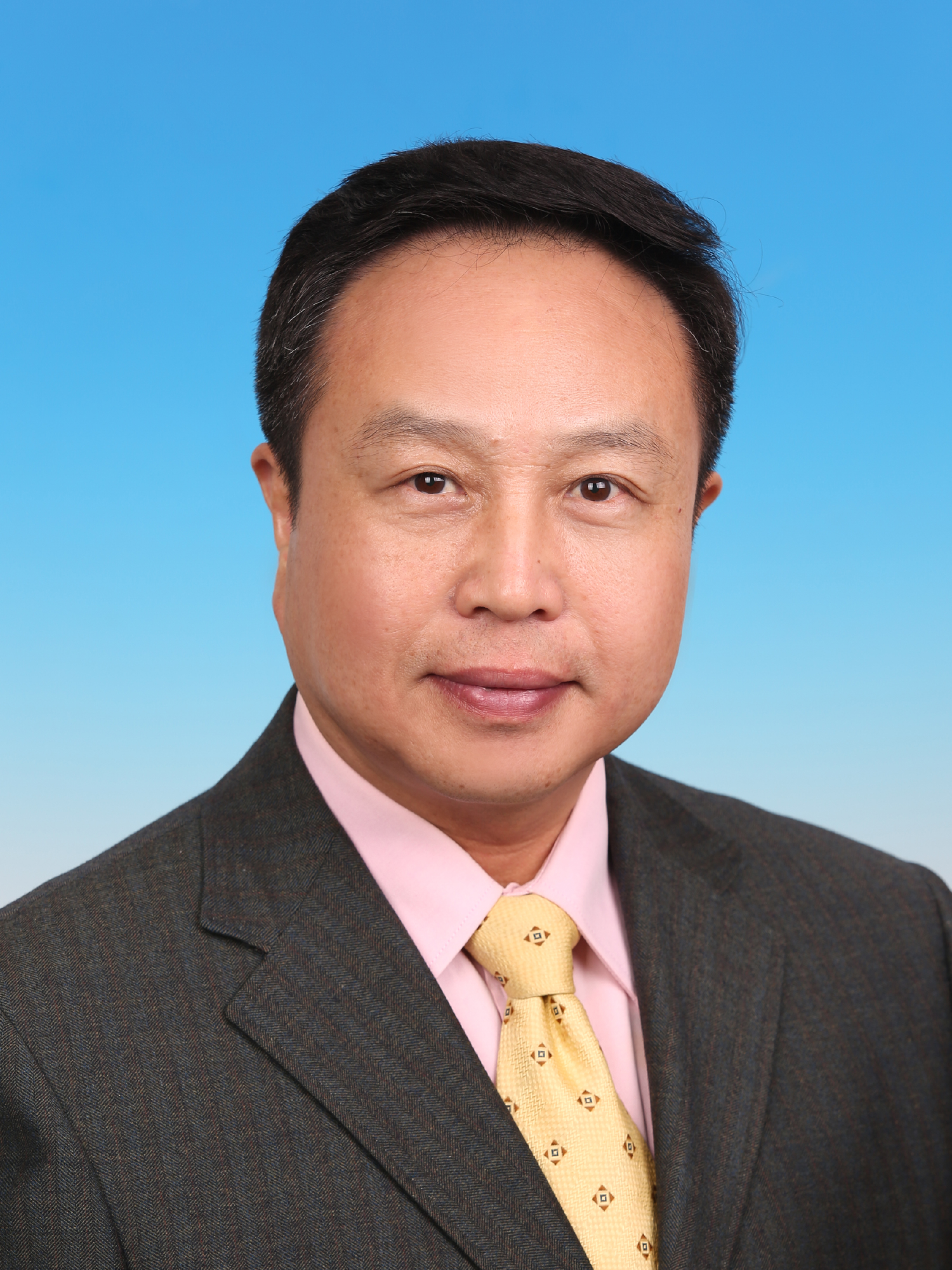}}]{Haibin Liu}
received the B.Eng. and the M.Eng. degrees in aeronautics and astronautics engineering from Northwestern Polytechnical University, Xi’an, China, in 1985 and 1988, respectively, and Ph.D. degree in precision engineering from Hokkaido University, Sapporo, Japan, in 1995. 
He was a Research Fellow with the Institute of NEC Foundation of Computer and Communication (C\&C) Promotion, Japan. From 1998 to 2004, he was a Principle Scientist with Motorola Japan Ltd., Japan. He was a CTO and Full Professor at the China Academy of Aerospace Systems Science and Engineering in Beijing, China. He is currently a Full Distinguished Professor with the College of Intelligent Machinery as a Deputy Director, Faculty of Materials and Manufacturing, Beijing University of Technology, Beijing, China. His current research interests include artificial intelligence, smart manufacturing, and hybrid control systems. 
\end{IEEEbiography}



\begin{IEEEbiography}[{\includegraphics[width=1in,height=1.25in,clip,keepaspectratio]{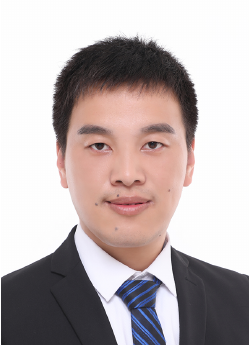}}]{Wenshuo Wang} (SM'15-M'18)
received Ph.D. degree in mechanical engineering from the Beijing Institute of Technology, Beijing, China, in 2018. 
Currently, he is a Postdoctoral Researcher at McGill University, Canada. Before joining McGill, he was a Postdoctoral Research Associate with Carnegie Mellon University and UC Berkeley from 2018 to 2020. He was also a Research Assistant with UC Berkeley and University of Michigan from 2015 to 2018. His research interests lie in Bayesian nonparametric learning, human driver model, human–vehicle interaction, ADAS, and autonomous vehicles.
\end{IEEEbiography}

\begin{IEEEbiography}[{\includegraphics[width=1in,height=1.25in,clip,keepaspectratio]{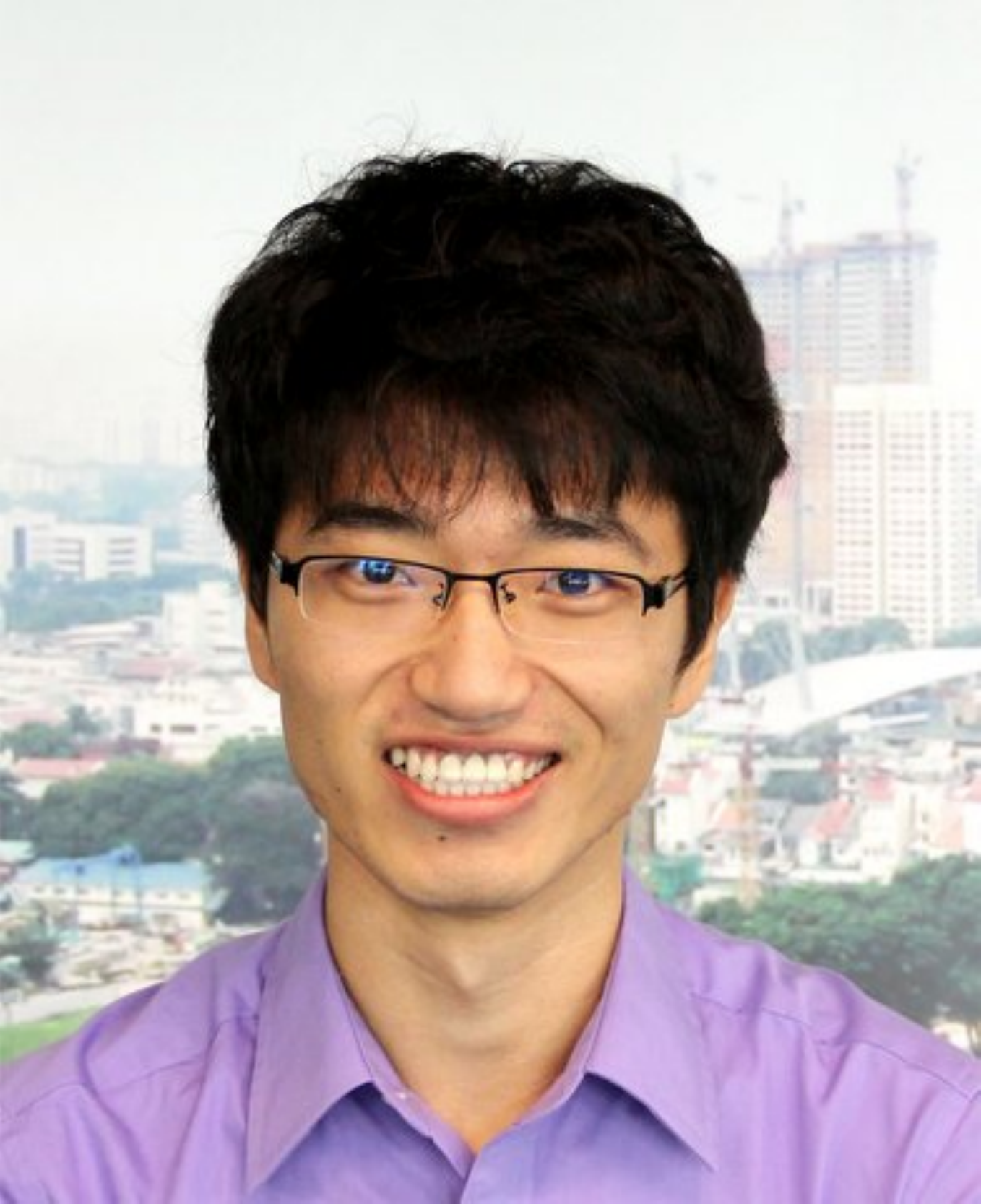}}]{Lijun Sun}
received B.S. degree in Civil Engineering from Tsinghua University, Beijing, China, in 2011 and Ph.D. degree in Civil Engineering (Transportation) from the National University of Singapore in 2015. He is currently an Assistant Professor with the Department of Civil Engineering at McGill University, Montreal, QC, Canada. His research centers on intelligent transportation systems, machine learning, spatiotemporal modeling, travel behavior, and agent-based simulation. He is an Associate Editor of Transportation Research Part C: Emerging Technologies. 
\end{IEEEbiography}

\vfill

\enlargethispage{-1in}






\end{document}